\RequirePackage{fix-cm}
\documentclass[smallextended]{svjour3}       
\smartqed  
\usepackage{amsmath,amssymb}
\usepackage{graphicx}
\usepackage{subfig}
\usepackage{tikz}
\usepackage{float}
\usepackage{hyperref}

\newcommand{\T}{\mathcal{P}}
\newcommand{\R}{\mathcal{R}}
\renewcommand{\S}{\mathcal{S}}
\newcommand{\A}{\mathcal{A}}

\newcommand\IfO{\mbox{{IfO}}}

\newcommand\RL{\mbox{\emph{RL}}}
\newcommand\BC{\mbox{\emph{BC}}}

\newcommand\IRL{\mbox{\emph{IRL}}}
\newcommand\GAIL{\mbox{\emph{GAIL}}}

\newcommand\GAN{\mbox{\emph{GAN}}}
\newcommand\MDP{\mbox{\emph{MDP}}}

\usepackage{natbib}
\definecolor{light-blue}{rgb}{.27, .87, .6}

\usepackage{fancyhdr}

\begin{document}

\title{Recent Advances in Leveraging Human Guidance for Sequential Decision-Making Tasks
}


\author{Ruohan Zhang \and Faraz Torabi \and Garrett Warnell\and Peter Stone
}


\institute{Ruohan Zhang$^{1*}$ \at
              \email{zharu@utexas.edu}\\
        Faraz Torabi$^{1*}$ \at
              \email{faraztrb@utexas.edu}\\
        Garrett Warnell$^{2}$  \at
              \email{warnellg@cs.utexas.edu}\\
        Peter Stone$^{1,3}$ \at
              \email{pstone@cs.utexas.edu}\\
        $^*$Contributed equally to this work\\
        $^1$Department of Computer Science, The University of Texas at Austin \\
        $^2$U.S. Army Research Laboratory \\
        $^3$Sony AI\\
}

\date{Accepted: 3 June 2021 | Published online: 23 June 2021}

\maketitle
\thispagestyle{fancy}

\begin{abstract}
A longstanding goal of artificial intelligence is to create artificial agents capable of learning to perform tasks that require sequential decision making.
Importantly, while it is the artificial agent that learns and acts, it is still up to humans to specify the particular task to be performed.
Classical task-specification approaches typically involve humans providing stationary reward functions or explicit demonstrations of the desired tasks.
However, there has recently been a great deal of research energy invested in exploring alternative ways in which humans may guide learning agents that may, e.g., be more suitable for certain tasks or require less human effort.
This survey provides a high-level overview of five recent machine learning frameworks that primarily rely on human guidance apart from pre-specified reward functions or conventional, step-by-step action demonstrations.
We review the motivation, assumptions, and implementation of each framework, and we discuss possible future research directions. 

\end{abstract}

\section{Introduction}
\label{intro}
Artificial agents require humans to specify the tasks they should perform.
With respect to artificial learning agents in particular, humans must provide some specification of what the agent should learn to perform.
One method by which humans typically provide this specification is by designing a stationary reward function.
This function provides a reward to the agent when it correctly performs the desired task and, perhaps, punishment when the agent does not.
Artificial learning agents may then approach the task-learning process using {\em reinforcement learning} (RL) techniques \citep{sutton2018reinforcement} that seek to find a {\em policy} (i.e., an explicit function that the agent uses to make decisions) that allows the agent to gather as much reward as possible.
Another popular way in which humans specify tasks for artificial agents to learn is by demonstrating the task themselves.
Typically, this is accomplished by having the human perform the task while the learning agent observes the actions that the human takes (e.g., the human physically moving a robot arm).
In these cases, artificial agents may use approaches from {\em imitation learning} (IL) \citep{schaal1999imitation,argall2009survey,osa2018algorithmic} in order to find policies that allow them to perform the demonstrated task.
Both paradigms described above (i.e., RL and IL) have been used with remarkable success \citep{mnih2015human,silver2016mastering,levine2016end,silver2017mastering,silver2018general,jaderberg2019human,vinyals2019grandmaster}, especially when combined with deep learning~\citep{lecun2015deep} to solve challenging sequential decision-making tasks.

While reward functions and explicit action demonstrations currently represent the most common ways in which humans specify tasks for learning agents, recent years have seen a great deal of research energy devoted to studying alternative ways in which humans might perform task specifications.
In general, these alternatives are focused on more diverse and creative ways of providing input than the two methods described above, and so we explicitly refer to the resulting types of input as {\em human guidance}.
Because human guidance is less direct compared to specified reward functions or explicit action demonstrations, attempts to leverage it have led to several new research challenges in the machine learning community.

There are many reasons for the recent interest in utilizing human guidance.
One reason is the relative ease with which several forms of human guidance can be collected.
For some tasks, it may be exceedingly difficult for a human trainer to specify a reward function or provide an action demonstration since both require some level of training and skill that the human may not possess.
However, it may still be possible for the human to guide the learning agent.
As an analogy from human learning, consider the sports coach that provides guidance in the form of feedback on professional athlete performance.
Even though the coach typically can not explicitly demonstrate the skill to be performed at the same skill or performance level as the athlete, their feedback is often useful to the athlete.
In these cases, the availability of guidance may even help the learner achieve greater final task performance than if an action demonstration alone was provided.
Another reason for the research community's interest in studying machine learning from human guidance lies in the utility of human guidance as a supplemental training signal that can increase the speed of task learning.
That is, even in cases for which a reward signal or an action demonstration is available, if the learning agent can leverage available human guidance, the overall amount of time it takes to arrive at an acceptable behavior policy can be greatly reduced compared to if the guidance had not been used at all.

This survey aims at providing a high-level overview of recent research efforts that primarily rely on human guidance as opposed to conventional reward functions or step-by-step action demonstrations.
We will define and discuss learning from five forms of human guidance~\citep{zhang2019leveraging}, including {\em (1)} evaluative feedback, {\em (2)} preferences, {\em (3)} high-level goals (hierarchical imitation), {\em (4)} demonstration sequences without actions (imitation from observation), and {\em (5)} attention. Though the approaches to be discussed vary with regards to the trade-off between the amount of information provided to the agent and the amount of human effort required, all have shown promising results in one or more challenging sequential decision-making tasks.

\section{Background}
\label{sec:bg}
In this section, we provide background relevant to the rest of the paper.
More specifically, we first discuss Markov decision processes (\MDP s), reinforcement learning, and the notation used in this paper.
We then provide a short review of imitation learning.

\subsection{Markov Decision Processes (\MDP s)}
\begin{figure}
    \centering
    \includegraphics[width=0.75\textwidth]{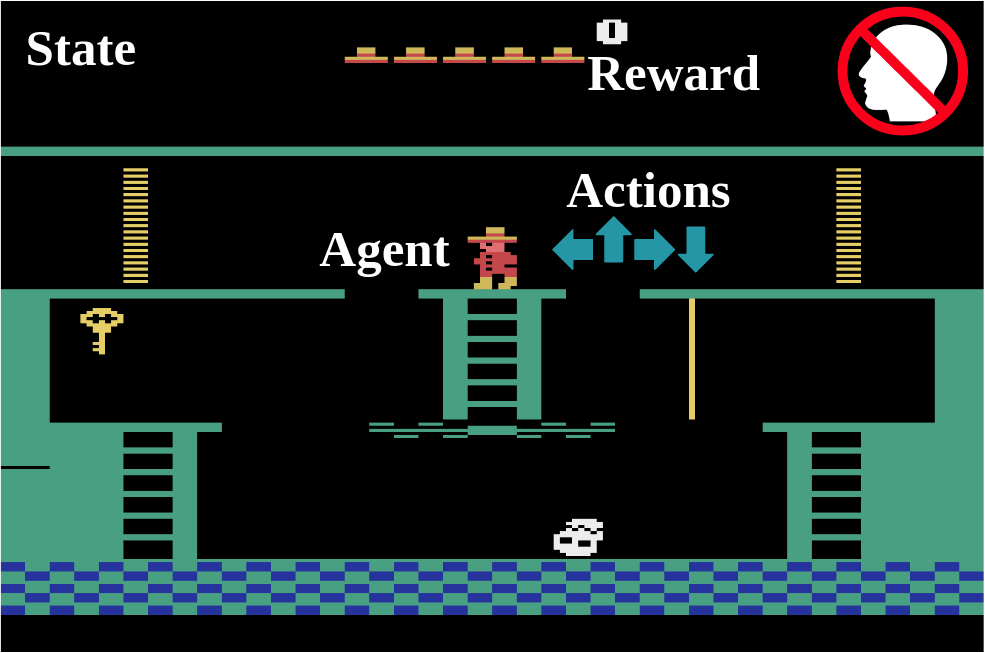}
    \caption{Atari Montezuma's Revenge is a challenging sequential decision-making task that is widely used in reinforcement learning research. The problem is modeled as a Markov decision process. 
    In a typical reinforcement learning setting, the agent needs to learn to play the game without any human guidance purely based on the score provided by the environment. }
    \label{fig:montezuma_rl}
\end{figure}

A standard reinforcement learning problem is formalized as a Markov decision process (\MDP), defined as a tuple $\langle\S, \A, \T, \R, \gamma\rangle$~\citep{sutton2018reinforcement}, where
\begin{itemize}
    \item $\S$ is a set of environment states which encodes relevant information for an agent's decision.
    \item $\A$ is a set of agent actions.
    \item  $\T$ is the state transition function which describes $p(s'|s,a)$, i.e., the probability of entering state $s'$ when an agent takes action $a$ in state $s$.
    \item $\R$ is a reward function. $r(s,a,s')$ denotes the scalar reward agent received on transition from $s$ to $s'$ under action $a$.
    \item  $\gamma \in \lbrack 0,1 \rbrack$ is a discount factor that indicates how much the agent values an immediate reward compared to a future reward.
\end{itemize}
As a concrete example, Atari Montezuma's Revenge~\citep{bellemare2013arcade} (Fig.~\ref{fig:montezuma_rl}) is one of the most challenging video games for reinforcement learning research. The game has rich visual features, complicated game dynamics, and very sparse rewards. Modeled as an MDP, the state is the game image frame, or a stack of frames to capture temporal information. The agent controls the avatar by choosing an action from a discrete set of actions at every timestep. The agent receives the reward from the game engine in the form of game scores. We will use this game as a running example throughout the survey.

Additionally, $\pi: \S \times \A \mapsto \lbrack 0,1 \rbrack$ is a policy which specifies the probability distribution of selecting actions in a given state. The goal for a learning agent is to find an optimal policy $\pi^*$ that maximizes the expected cumulative reward. One could optimize $\pi$ directly, while alternatively many of the algorithms are based on value function estimation, i.e., estimating the state value function $V^{\pi}(s)$ or the action-value function $Q^{\pi}(s,a)$. 

The state value function for a given policy $\pi$ is defined as~\citep{sutton2018reinforcement}
\begin{equation}
V^\pi(s) = \mathbb{E}_\pi \left[ \sum_{t=0} \gamma^t R(s_t,a_t) \; | \; s_0 = s \right]
\label{eq:v_value}
\end{equation}
A corresponding action value function, $Q^\pi(s,a)$, also exists and is given by 
\begin{equation}
Q^\pi(s,a) = E_\pi \left[ R(s_t,a_t) + V^\pi(s_{t+1}) \; | \; s_t=s, a_t=a \right]
\label{eq:q_value}
\end{equation}
and the advantage function $A^\pi(s,a)$, is defined as 
\begin{equation}
A^\pi(s,a) = Q^\pi(s,a) - V^\pi(s).
\label{eq:a_value}
\end{equation}
The state value function $V^\pi(s)$ measures the expected cumulative reward to be in a particular state $s$ and following policy $\pi$ afterward. The action-value function $Q^\pi(s,a)$ defines the same quantity but for taking a particular action $a$ when in state $s$ and following policy $\pi$ afterward. The advantage function tells us the relative gain (``advantage") that could be obtained by taking a certain action compared to the average action taken at that state~\citep{wang2016dueling}.

Several successful RL algorithms that seek to estimate these quantities directly have been developed, including Q Learning \citep{watkins1992q} and advantage actor-critic (see, e.g., \cite{sutton2018reinforcement}).
For example, Q-learning seeks to learn the state-action value function for the optimal policy, $Q^{\pi^*}(s,a)$, and the policy is then given by $\pi^*(s) = \arg \max_a Q^{\pi^*}(s,a)$. Nowadays, deep neural networks are often used as function approximators to estimate and optimize $\pi$, $V$, and $Q$.

An important challenge in RL is to balance exploration vs. exploitation when an agent selects its action. Exploration allows the agent to improve its current knowledge. Exploitation chooses the greedy action to maximize reward by exploiting the agent’s current knowledge. A simple strategy ($\epsilon$-greedy) chooses a random action with probability $\epsilon$ and chooses the greedy action (the action with the highest Q value) with probability $1-\epsilon$~\citep{sutton2018reinforcement}. A more sophisticated strategy uses a Boltzmann distribution for selecting actions based on the current estimate of Q function~\citep{sutton2018reinforcement}:
\begin{equation}
\label{eqn:softmax}
P(a|s, Q, \tau) = \frac{e^{Q(s,a)}/\tau}{\sum_{a' \in \A} e^{Q(s,a')}/\tau}
\end{equation}
where $\tau$ is a temperature constant that controls the exploration rate.

\subsection{Imitation Learning}
\label{sec:IL}
\begin{figure}
    \centering
    \includegraphics[width=1\textwidth]{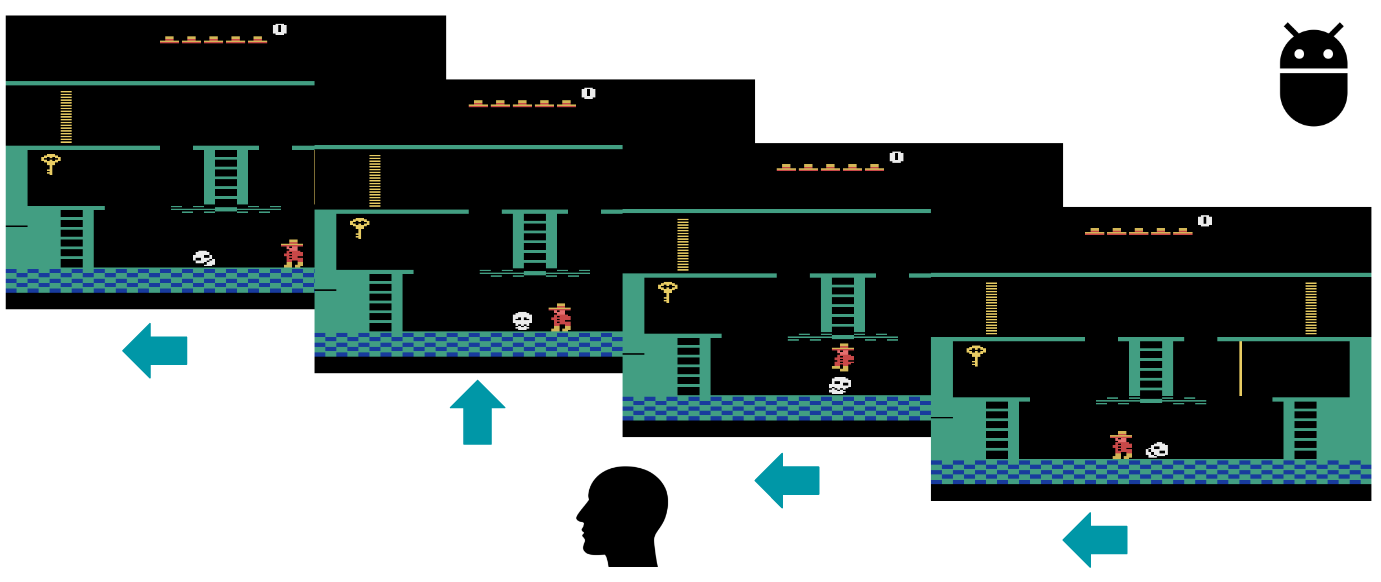}
    \caption{In standard imitation learning, a human trainer demonstrates a sequence of actions, and the agent learns to imitate the trainer's actions using behavioral cloning, inverse reinforcement learning, or adversarial imitation.}
    \label{fig:mtzm_il}
\end{figure}

The learning frameworks surveyed in this paper are inspired by, an extension of, or combined with traditional imitation learning algorithms. The standard imitation learning setting (Fig.~\ref{fig:mtzm_il} and Fig.~\ref{fig:il}) can be formulated as MDP\textbackslash$R$, i.e. there is no reward function $\R$ available. Instead, a learning agent (the \emph{imitator}) records expert (the \emph{demonstrator}, could be a human expert or an artificial agent) demonstrations in the format of state-action pairs $\{(s_t,a^*_t)\}$ at each timestep, and then attempts to learn the task using that data. 

One approach is for the agent to learn to mimic the demonstrated policy using supervised learning, which is known as behavioral cloning~\citep{bain1999framework}.
A second approach to imitation learning is called inverse reinforcement learning (\IRL) \citep{abbeel2004apprenticeship} which involves learning a reward function based on the demonstration data and learning the imitation policy using RL with the learned reward function.
These two approaches constitute the major learning frameworks used in imitation learning. Comprehensive reviews of these two approaches can be found in~\cite{argall2009survey,hussein2017imitation,osa2018algorithmic,arora2018survey,fang2019survey}.
More recently, generative adversarial imitation learning (\GAIL) \citep{ho2016generative} has been proposed, which utilizes the notion of generative adversarial networks (\GAN) \citep{goodfellow2014generative}.

Importantly, all of these approaches assume that $(s_t,a^*_t)$ pairs are the only learning signal to the agent and that both $s_t$ and $a^*_t$ are available to the agent.
Unfortunately, access to the optimal actions, $a^*_t$, is not always plausible, as the task might be too complex for human demonstrators to perform.
Therefore, there has recently been a great deal of interest in the research community in learning frameworks that utilize learning signals other than optimal action information, and it is these techniques that we review in this survey.
Before doing so, however, we briefly describe the three IL frameworks described above, i.e., {\em (1)} behavioral cloning (\BC), {\em (2)} inverse reinforcement learning (\IRL), and {\em (3)} generative adversarial imitation learning (\GAIL).



\subsubsection{Behavioral Cloning (\BC)}
Behavioral cloning \citep{pomerleau1989alvinn,bain1999framework} is one of the main methods to approach an imitation learning problem.
The agent receives as training data both the encountered states and actions of the demonstrator, then uses supervised learning techniques such as classification or regression to estimate the demonstrator's policy.
This method is powerful in the sense that it is capable of imitating the demonstrator without having to interact with the environment, and it has been successfully applied in many application domains.
For instance, it has been used to train a quadrotor to fly down a forest trail \citep{giusti2016machine}.
There, the training data consists of images of the forest trail gathered by cameras mounted on a human hiker and labeled with the actions (walking directions) that the human used.
The policy is modeled as a convolutional neural network classifier, and trained using supervised learning.
In the end, the quadrotor managed to fly down the trail successfully.
BC has also been used in autonomous driving \citep{bojarski2016end}.
The training data is acquired using a human demonstrator, and a convolutional neural network is trained to map raw pixels from a single front-facing camera directly to platform steering commands.
After training, the vehicle was capable of driving in traffic on local roads.
BC has also been successfully used to teach robotic manipulators complex, multi-step, real-world tasks using kinesthetic demonstrations \citep{niekum2015learning}.

One of BC's major drawbacks is potential performance degradation due to the well-studied compounding error caused by covariate shift \citep{ross2010efficient,ross2011reduction}, i.e., that training and testing data distribution mismatch results in deviation of the learned behavior from the demonstration \citep{torabi2018behavioral}.
\cite{ross2011reduction} proposed an interactive training method to correct the shift called DAgger (Dataset Aggregation) which attempts to bring the distribution of demonstration data closer to that of the learned behavior. It does so by collecting demonstration data on the states observed by the imitator at each iteration. Retraining the policy on the aggregated dataset ultimately prevents the imitator from deviating from the demonstration behavior.

\subsubsection{Inverse Reinforcement Learning (\IRL)}
Inverse reinforcement learning \citep{abbeel2004apprenticeship,ziebart2008maximum} is a second category of imitation learning. IRL  techniques seek to learn a reward function that has the maximum value for the demonstrated actions.
The learned reward function is then used in combination with RL methods to find an imitation policy. To be more specific, most IRL algorithms first initialize a random policy. Next, the agent executes that policy in the environment to collect state-action data, and then the algorithms estimate the expert's reward function based on the data generated by the policy and the demonstration data. Finally, standard RL algorithms are used to learn an optimal policy for that reward function. The process of reward learning and policy learning is repeated until the agent policy becomes sufficiently close to the demonstrator's policy. 
Like BC techniques, IRL methods usually assume that state-action pairs are available \citep{finn2016guided}, and also that the reward is a function of both states and actions. The algorithms developed in this category have shown impressive results in a variety of tasks such as autonomous helicopter aerobatics \citep{abbeel2010autonomous}, robot object manipulation \citep{finn2016guided}, and autonomous navigation in complex unstructured terrains \citep{silver2010learning}, etc.

One major drawback of most algorithms developed for IRL is that at each iteration, they have to solve a complete RL problem to find an optimal policy given the currently estimated reward function which is computationally very expensive. However, the learned policies are often more robust than the policies learned by BC algorithms as they do not suffer from the covariate shift problem. This shift does not happen in the case of IRL because the agent can interact with the environment while training and the distribution mismatch diminishes during the process.

\subsubsection{Adversarial Imitation Learning}
Recently an imitation learning algorithm, generative adversarial imitation learning (\GAIL) \citep{ho2016generative}, has been developed that alleviates the IRL's drawback just set forth. This algorithm directly learns the policy given demonstration bypassing the optimal reward recovery. \GAIL ~formulates the problem of finding an imitating policy as that of solving the following optimization problem:
\begin{equation}\label{gail}
\begin{split}
\min_{\pi \in \prod} \displaystyle{\max_{D \in (0,1)^{\mathcal{S} \times \mathcal{A}}}} & -\lambda_H H(\pi) + \mathbb{E}_\pi[\log(D(s,a)] +\\ &\mathbb{E}_{\pi_E}[\log(1-D(s,a))]\;,
\end{split}
\end{equation}
where $\prod$ is the set of all stationary stochastic policies, $\pi_E$ is the demonstrator's policy, $\lambda_H$ is a weight factor, $H$ is the entropy function, and the discriminator function $D:\mathcal{S} \times \mathcal{A} \rightarrow (0,1)$ can be thought of as a classifier trained to differentiate between the state-action pairs provided by the demonstrator and those experienced by the imitator.
The objective in (\ref{gail}) is inspired by the one used in generative adversarial networks (\GAN s) \citep{goodfellow2014generative}. A \GAN~system is trained in a competitive process: the generator tries to fool the classifier while the classifier tries to distinguish the generated data from the real data. This competitive training process makes both models do better by trying to beat the other. In \GAIL~the associated algorithm can be thought of as trying to induce an imitator state-action occupancy measure that is similar to that of the demonstrator. $\pi$ and $D$ are often parameterized in practice and that \GAIL seeks to find the saddle point of Eq.~\ref{gail} by sequentially making gradient steps with respect to the parametrization of $D$ and $\pi$. Trust Region Policy Optimization (TRPO)~\citep{schulman2015trust} is often used to update the policy. Maximizing the entropy term $H(\pi)$ follows the maximum causal entropy IRL~\citep{ziebart2008maximum,ziebart2010modeling,bloem2014infinite}. The entropy serves as a policy regularizer to account for the noise and suboptimality in the demonstrated behavior~\citep{ziebart2008maximum}. Even more recently, there has been research on methods that seek to improve on \GAIL ~by, e.g., increasing sample efficiency \citep{kostrikov2018discriminatoractorcritic,sasaki2018sample} and improving reward representation \citep{fu2018learning,qureshi2018adversarial}.

\subsection{Common Task Domains}
Next, we introduce several sequential decision-making tasks that are commonly used today to test the algorithms discussed above. Before deep learning, sequential decision-making models and learning algorithms were often confined to task domains with low dimensional state space such as 2D gridworld and mountain car~\citep{sutton1998reinforcement}. The emergence of deep neural networks~\citep{lecun2015deep} has enabled these models and algorithms to solve significantly more challenging tasks. These tasks include Atari 2600 video games~\citep{bellemare2013arcade,machado2018revisiting} in which the state space could be high-dimensional raw game images. The platform has 60 unique video games that span a variety of dynamics, visual features, reward mechanisms, and difficulty levels for both humans and AIs. Montezuma's Revenge (Fig.~\ref{fig:montezuma_rl}) is one of the most difficult games due to very sparse rewards. Hence it is one of the most challenging games in terms of exploration.

Another example is robotic locomotion tasks using MuJoCo~\citep{todorov2012mujoco}, a simulator with a physics engine and multi-joint dynamics to study complex dynamical systems in contact-rich behaviors. It is the first full-featured simulator designed from the ground up for the purpose of model-based optimization, and in particular optimization through contacts~\citep{todorov2012mujoco}.

\begin{figure*}
\centering
\subfloat[Atari Bowling]{\includegraphics[width=0.33\textwidth]{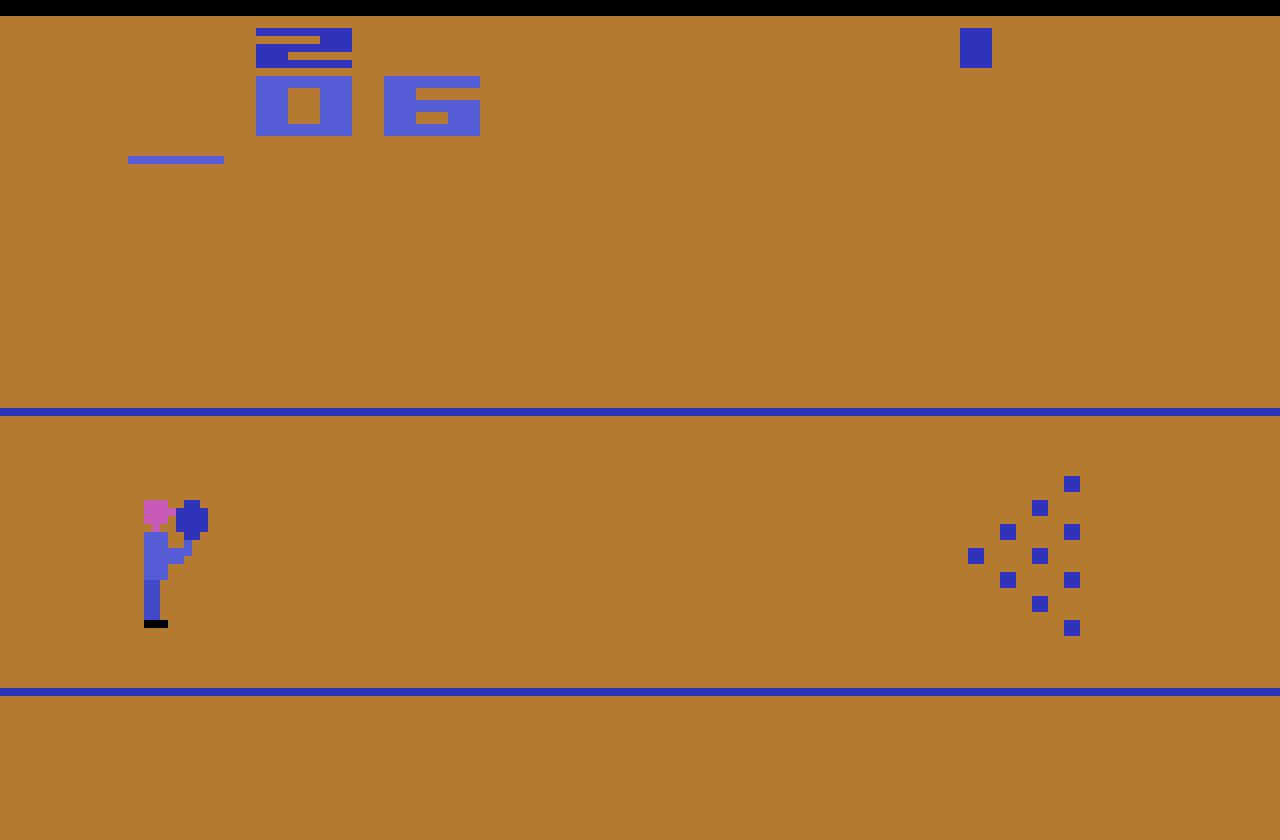}}
\hspace{0.005cm}
\subfloat[Atari Montezuma's Revenge]{\includegraphics[width=0.33\textwidth]{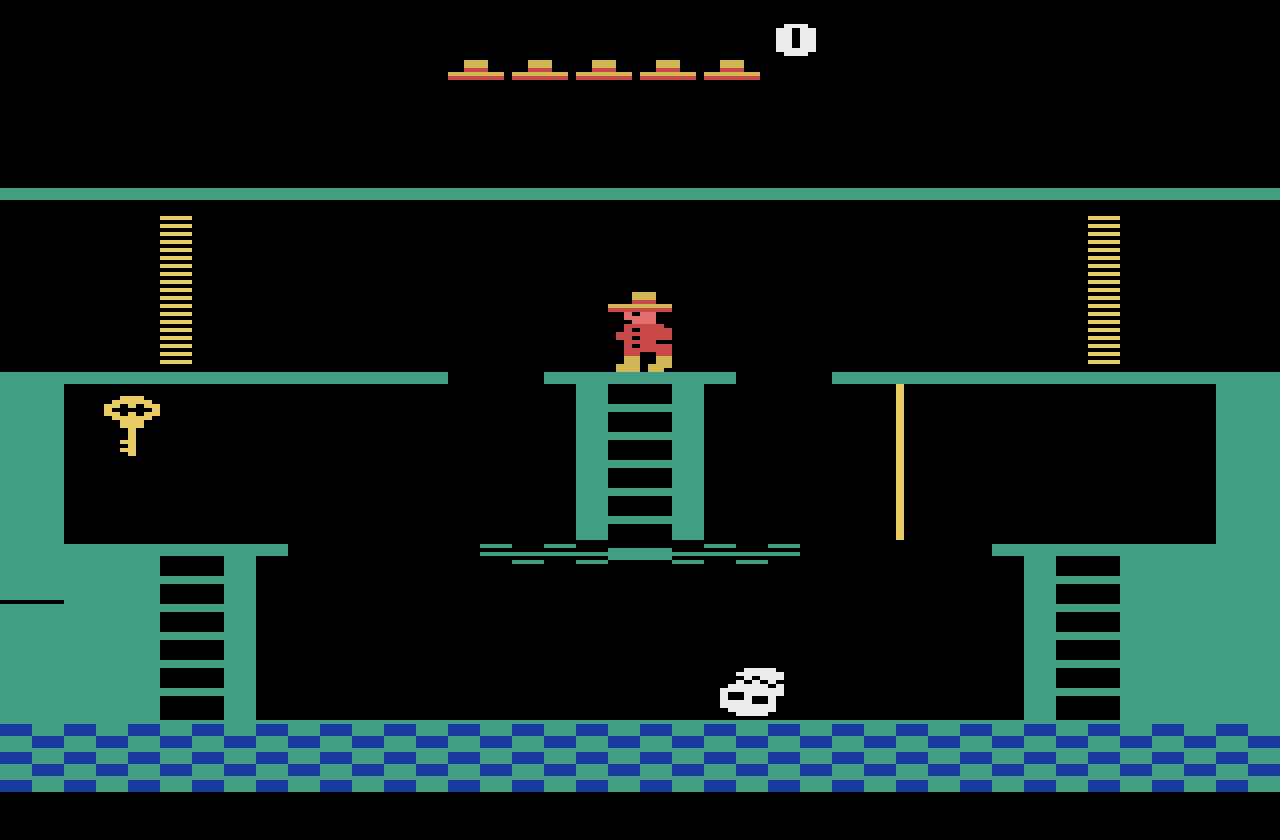}}
\hspace{0.005cm}
\subfloat[MuJoCo Ant]{\includegraphics[width=0.2725\textwidth]{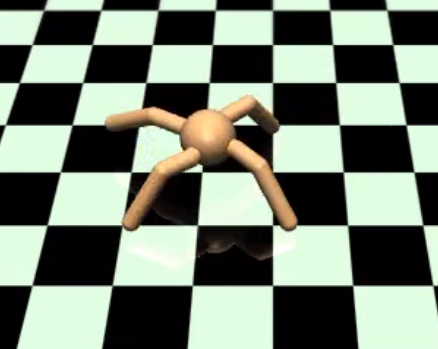}}
\\
\subfloat[Using TurtleBot for navigation and human-robot interaction tasks, adapted from \cite{macglashan2017interactive}]{\includegraphics[width=0.35\textwidth]{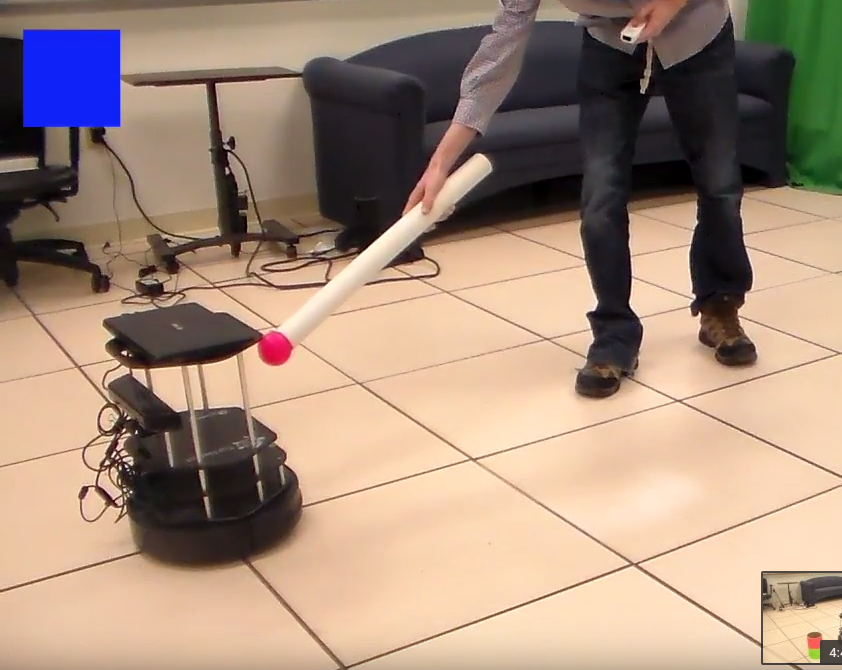}}
\hspace{0.005cm}
\subfloat[Simulated and real robot manipulation (table clean-up), adapted from \cite{xu2018neural}]{\includegraphics[width=0.5875\textwidth]{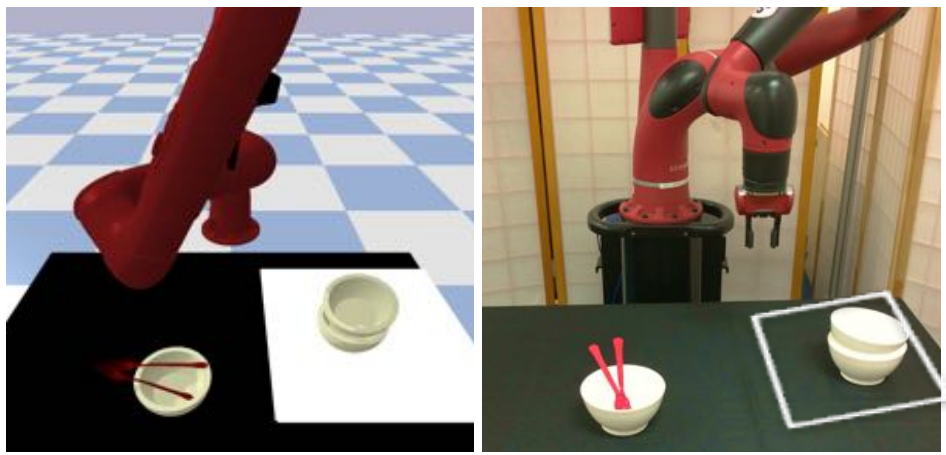}}
\\
\subfloat[Autonomous driving, adapted from~\cite{yu2018bdd100k}]{\includegraphics[width=0.75\textwidth]{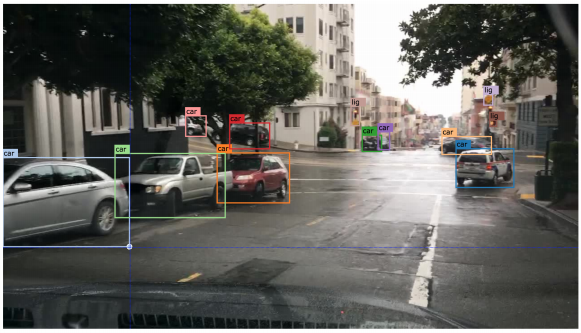}}
\caption{Example sequential decision tasks that are commonly used in recent research that leverages human guidance. The state spaces in these tasks are typically high-dimensional, such as raw images and robot joints with multiple degrees of freedom. }
\label{fig:tasks}
\end{figure*}

Recently, much effort has been spent on moving from simulation to real-world applications, from the navigation robots (e.g., TurtleBot~\citep{macglashan2017interactive}), robotic manipulators (e.g., Sawyer robot arm~\citep{xu2018neural}), to autonomous driving vehicles~\citep{yu2018bdd100k}. These are typically tasks with high-dimensional state space at which humans are particularly good. Examples of the tasks can be seen in Fig.~\ref{fig:tasks}. In some of the tasks such as board games, reinforcement learning agents have already surpassed human expert performance~\citep{silver2016mastering}, and could perform even better without human knowledge~\citep{silver2017mastering,silver2018general}. For example, \cite{silver2017mastering} have shown that a pure RL agent that learns to play Go by itself from scratch can outperform an IL/RL hybrid agent~\citep{silver2016mastering} which first learns to imitate expert human Go players' moves. However, RL agents and algorithms still face significant challenges in solving many of the tasks we discuss in this survey.

\section{Overview}
\begin{figure}[H]
\centering
\subfloat[Standard imitation learning]{\includegraphics[width=0.4\textwidth]{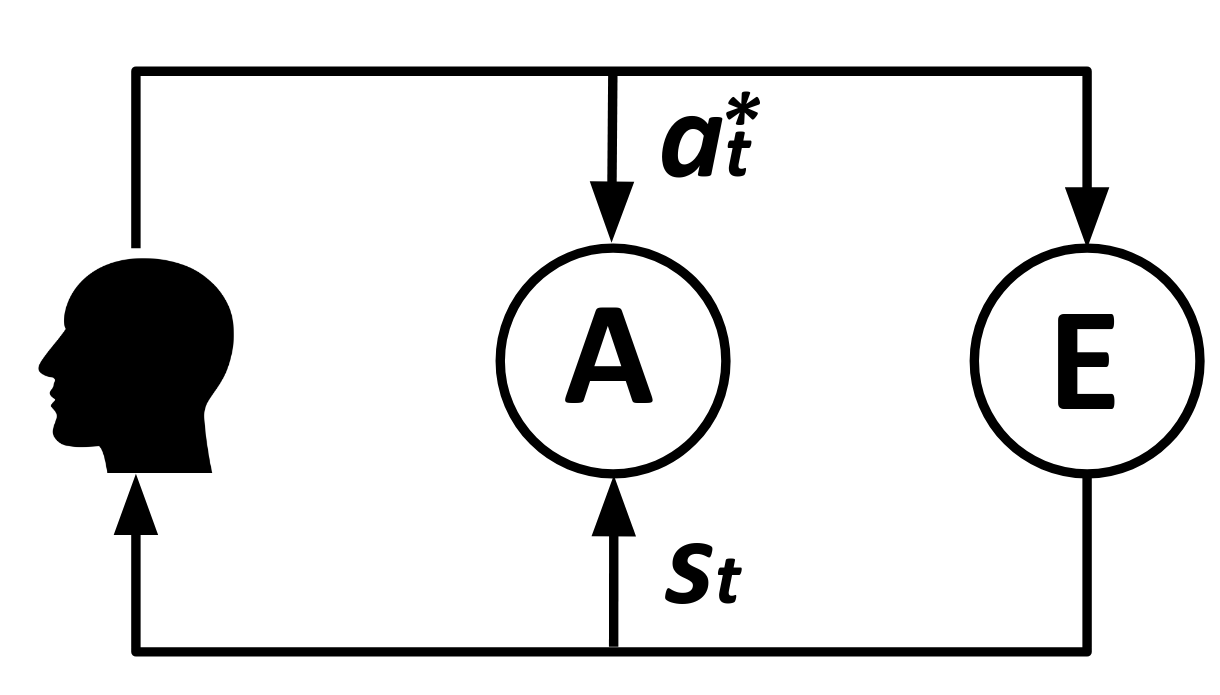}\label{fig:il}}
\subfloat[Evaluative feedback]{\includegraphics[width=0.4\textwidth]{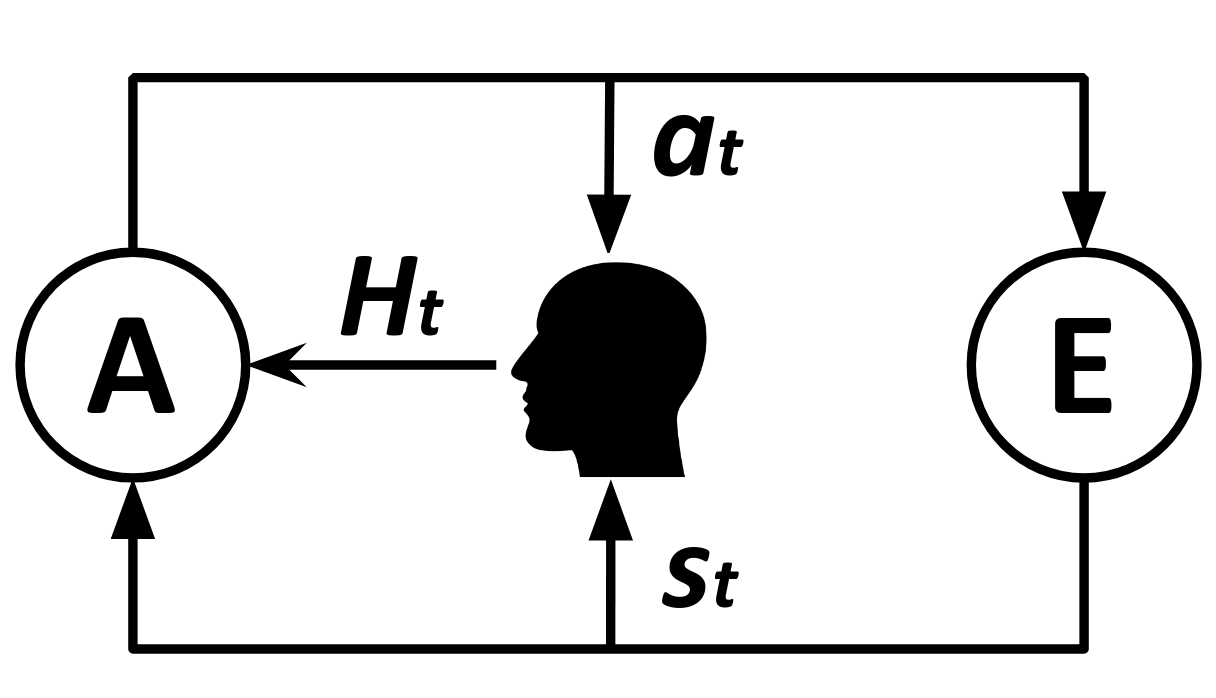}\label{fig:fb}}\\
\subfloat[Learning from human preference]{\includegraphics[width=0.4\textwidth]{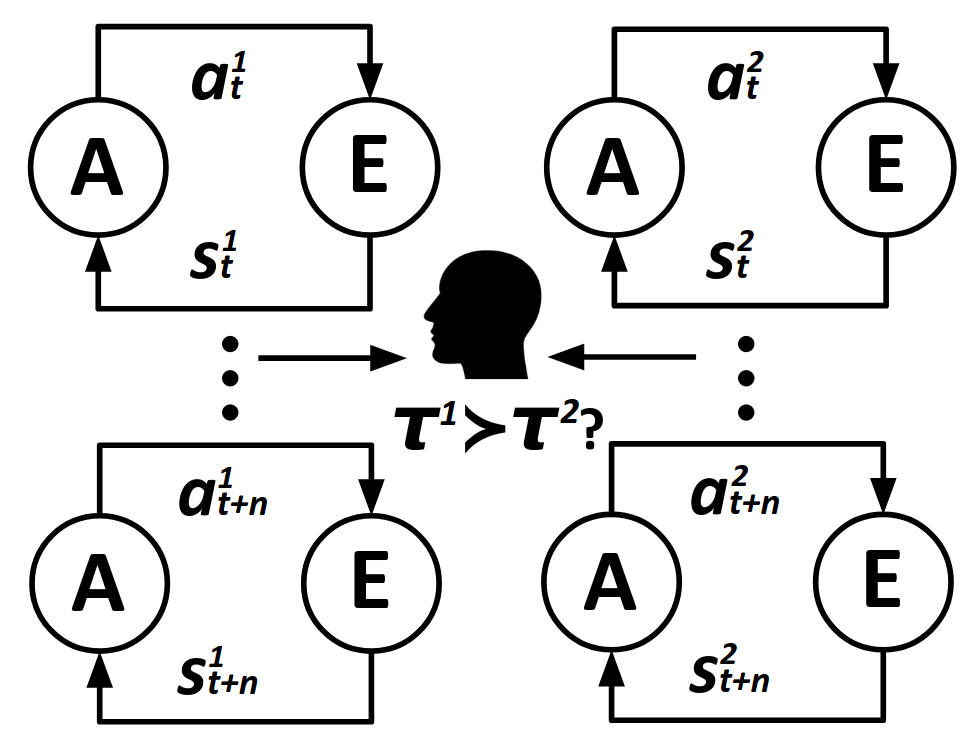}\label{fig:pref}}
\subfloat[Hierarchical imitation]{\includegraphics[width=0.4\textwidth]{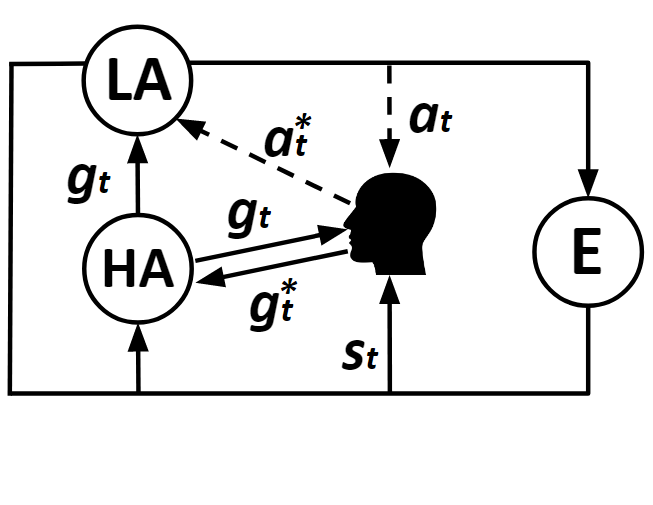}\label{fig:hi}}\\
\subfloat[Imitation from observation]{\includegraphics[width=0.4\textwidth]{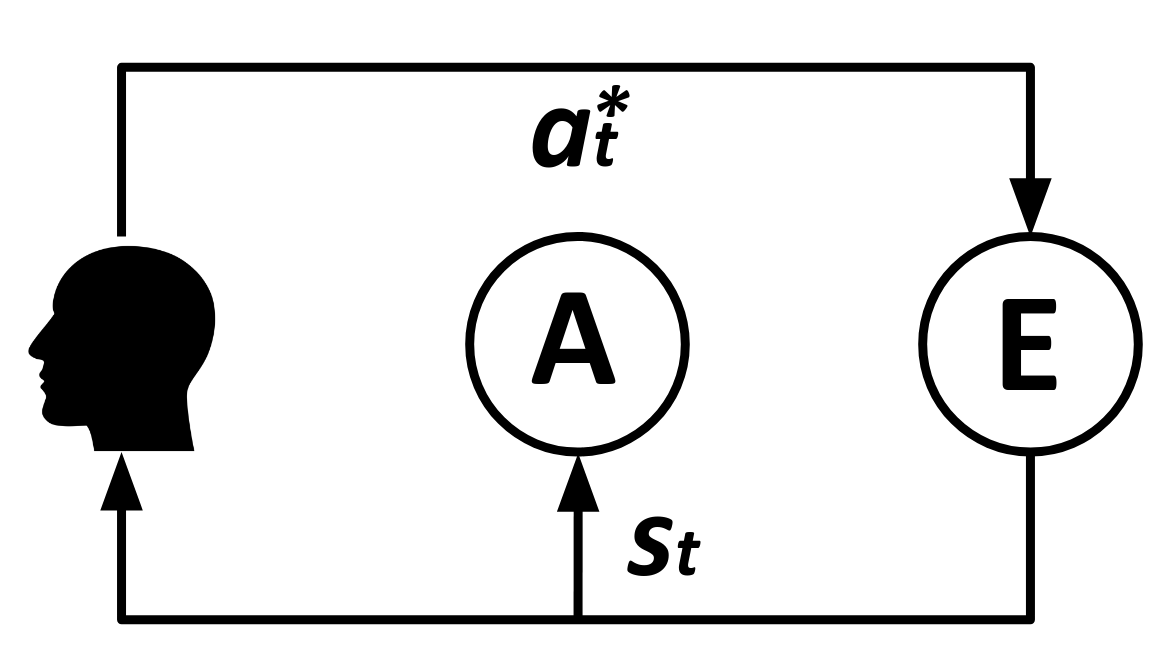}\label{fig:ifo}}
\subfloat[Learning attention from human]{\includegraphics[width=0.4\textwidth]{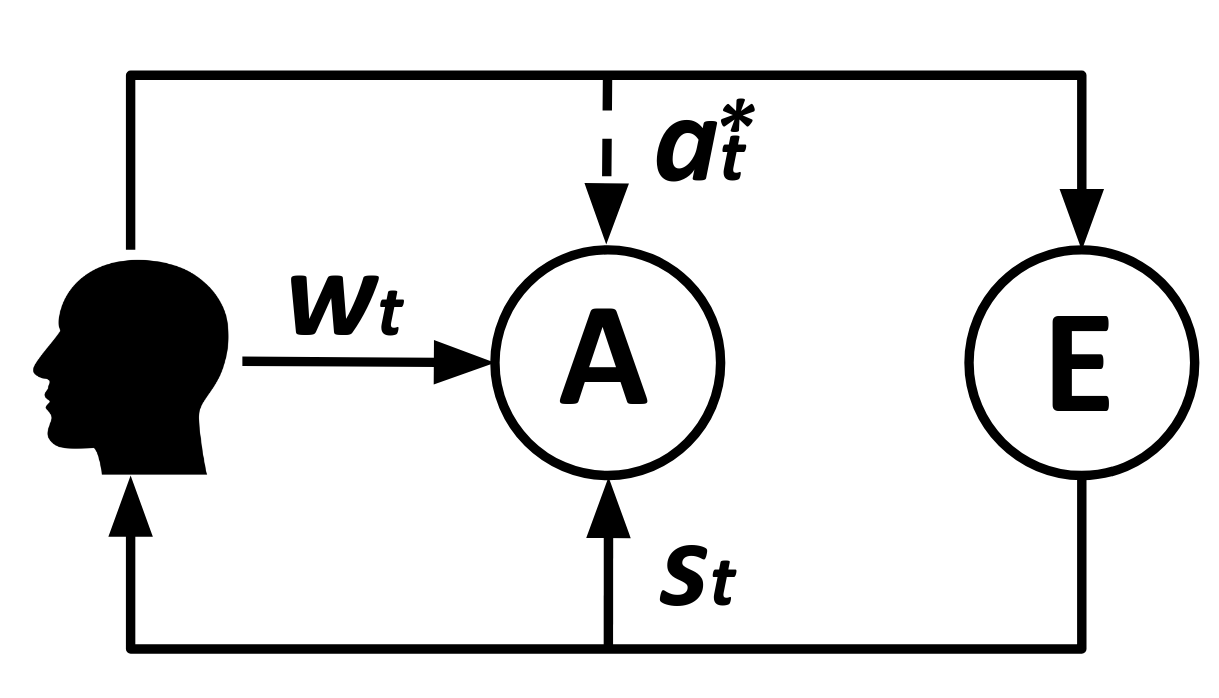}\label{fig:att}}\\
\caption{Human-agent-environment interaction diagrams of five learning frameworks surveyed. These diagrams illustrate how different types of human guidance data are collected, including information required by the human trainer and the guidance provided to the agent. Note that the learning process of the agent is not included in these diagrams. Arrow: information flow direction; Dashed arrow: optional information flow. \textbf{A}: learning agent; \textbf{E}: environment; $s_t$: the state at time $t$; $a_t$: agent action. (a) $a^*_t$: human demonstrated action. (b) $H_t$: human evaluative feedback on agent decision $a_t$ in state $s_t$. (c) $\tau^1 \succ \tau^2$: human trainer prefers agent behavior trajectory $\tau^1$ over $\tau^2$. (d) \textbf{HA}: a high-level agent that chooses a high-level goal $g_t$ for state $s_t$; \textbf{LA}: a low-level agent that chooses an action $a_t$ based on $g_t$ and $s_t$; $g^*_t$: high-level goal provided by human.
(e) Note that the human demonstrated action $a^*_t$ is not available to the agent. (f) $w_t$: human attention information.}
\label{fig:approach}
\end{figure}

Given the models and notations defined above, we now provide formal definitions of the five learning frameworks surveyed that leverage human guidance. Diagrams that visualize the interactions between the human trainers, the learning agents, and the task environment for imitation learning together with these five learning frameworks can be found in Fig.~\ref{fig:approach}.  In (a) standard imitation learning, the human trainer observes state information $s_t$ and demonstrates action $a^*_t$ to the agent; the agent stores this data to be used in learning later. In (b) learning from evaluative feedback, the human trainer does not perform the task, instead, he or she watches the agent performing the task, and provides instant feedback $H_t$ on agent decision $a_t$ in state $s_t$. In (c) learning from human preference. The human trainer watches two behaviors generated by the learning agent simultaneously and decides which behavior is more preferable. In (d) hierarchical imitation, The high-level agent chooses a high-level goal $g_t$ for state $s_t$. The low-level agent then chooses an action $a_t$ based on $g_t$ and $s_t$. The primary guidance that the trainer provides in this framework is the correct high-level goal $g^*_t$. Imitation from observation (e) is similar to standard imitation learning except that the agent does not have access to human demonstrated action -- it only observes the state sequence demonstrated by the human. Learning attention from humans (f) requires the trainer to provide attention information $w_t$ that indicates important task features to the learning agent. For each learning framework, a summary and comparison of selected papers surveyed can be found in Table~\ref{tbl:compare}. 

\begin{table*}
\scalebox{0.85}{
\centering
\begin{tabular}{c | c c c c c c}
\hline
Paper       & Human guidance       & Task domain(s)    & On-line?  & Dataset? & Section   \\
\hline
\cite{cederborg2015policy} & evaluative feedback & Pac-Man & Yes & No & 4 \\
\cite{warnell2018deep} & evaluative feedback & Atari Bowling & Yes & No & 4 \\
\cite{arumugam2019deep} & evaluative feedback & Minecraft & Yes & No & 4 \\
\cite{saunders2018trial} & evaluative feedback, actions & 3 Atari games & Yes & No & 4 \\
\cite{akinola2020accelerated} & evaluative feedback & simulated robot navigation & Yes & No & 4 \\
\hline
\cite{christiano2017deep} & preference & 8 MuJoCo tasks, 7 Atari games & Yes & No & 5\\
\cite{sadigh2017active} & preference & simulated driving & Yes & No & 5 \\
\cite{ibarz2018reward} & preference, actions & 9 Atari games & Yes & No & 5 \\ 
\cite{bestick2018learning} & preference & simulated and physical robot handover & Yes & No & 5 \\
\cite{cui2018active} & preference & simulated robot manipulation & Yes & No & 5 \\
\cite{palan2019learning} & preference, action & simulated driving, Lunar Lander, & Yes & No & 5 \\
& & simulated and physical robot manipulation & & & \\
\hline
\cite{le2018hierarchical} & high-level and low-level actions & Atari Montezuma's Revenge, maze navigation & Yes & No & 6 \\
\cite{andreas2017modular} & high-level actions & crafting, maze navigation, MuJoCo & No & No & 6 \\ 
\cite{gupta2020relay} & low-level actions & simulated robot manipulation & No & No & 6 \\
\cite{krishnan2017ddco} & low-level actions & robot manipulation & No & No & 6\\
\cite{codevilla2018end} & high-level and low-level actions & driving & No & \href{https://github.com/carla-simulator/imitation-learning}{Link} & 6 \\
\cite{xu2018neural} & high-level and low-level actions & robot manipulation & No & No &6 \\
\cite{fox2019multi} & high-level and low-level actions & robot manipulation & No & \href{https://github.com/BerkeleyAutomation/HIL-MT}{Link} & 6\\
\hline
\cite{torabi2018behavioral} & state & MuJoCo & No & No & 7\\
\cite{liu2018imitation} & state & MuJoCo, physical robot manipulation & No & No & 7\\
\cite{sermanet2018time} & state & physical robot manipulation & No & \href{https://sermanet.github.io/imitate/}{Link} & 7\\
\cite{torabi2018generative} & state & MuJoCo & No & No & 7\\
\cite{yang2019imitation} & state & MuJoCo & No & No & 7\\

\hline
\cite{palazzi2018predicting} & gaze & driving & No & \href{https://aimagelab.ing.unimore.it/imagelab/page.asp?IdPage=8}{Link} & 8\\
\cite{deng2019drivers} & gaze & driving & No & \href{https://github.com/taodeng/CDNN-traffic-saliency}{Link} & 8 \\
\cite{liu2019gaze} & gaze, action & driving & No & No & 8\\
\cite{xia2020periphery} & gaze, action & driving & No & \href{https://github.com/pascalxia/driver_attention_prediction}{Link} & 8 \\
\cite{zuo2018gaze} & gaze, action & non-verbal interaction & No & No & 8 \\ 
\cite{li2018eye} & gaze, action & meal preparation & No & \href{http://cbs.ic.gatech.edu/fpv/}{Link} & 8\\
\cite{zhang2020atari} & gaze, action & 20 Atari games & No & \href{https://zenodo.org/record/3451402}{Link} & 8\\
\hline
\hline
\end{tabular}
}
\caption{A comparison of selected papers surveyed. This table only includes recent works that aimed to solve task domains with high-dimensional state space. The ``On-line" column specifies whether the learning is done on-line or off-line, where on-line means that a human trainer must be available during the agent's learning process. ``Dataset" indicates whether associated human guidance data is published. If so the link to the dataset is provided. }
\label{tbl:compare}
\end{table*}

\section{Learning from Evaluative Feedback}
\begin{figure}
    \centering
    \includegraphics[width=1\textwidth]{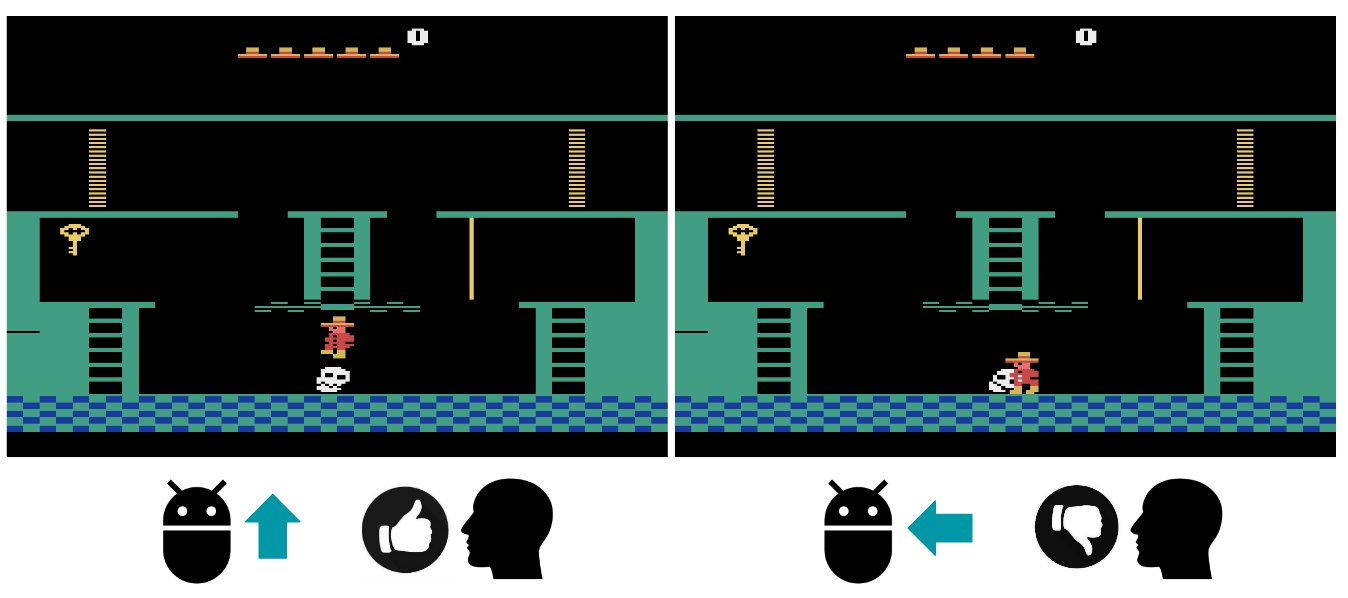}
    \caption{In learning from evaluative feedback, a human trainer watches the agent's learning process, and provides positive feedback for a desirable action (jumping over the skull), and negative feedback for an undesirable action (running into the skull).}
    \label{fig:mtzm_eval}
\end{figure}

We begin with one of the most natural forms of human guidance that have been studied: {\em evaluative feedback}. Proposed paradigms for learning from evaluative feedback typically involve human trainers watching artificial agents attempt to execute tasks and those humans providing a scalar signal that communicates the desirability of the observed agent behavior, as shown in Fig.~\ref{fig:fb} and \ref{fig:mtzm_eval}. Using this type of human guidance, the learning problem for the agent is that of determining how to adjust its policy such that its future behavior becomes more desirable to the human.

Evaluative feedback is an attractive form of human guidance due to the relative ease with which humans can provide it.
For example, for cases in which the human trainer cannot provide a demonstration of the task (because, e.g., the task is too difficult), the human typically still knows what constitutes good behavior and can therefore provide evaluative feedback.
Moreover, even when the human can provide a demonstration, providing additional evaluative feedback during the learning process may allow the artificial agent to achieve a task performance that exceeds that of the human demonstrator.

One of the main challenges faced by machines that seek to learn from human-provided evaluative feedback is that of correctly interpreting the feedback signal.
Indeed, several interpretations have been proposed by members of the research community, each leading to a different type of machine learning method.
Typically, the particular feedback interpretation manifests as equating the feedback with a particular quantity derived from the RL setting.
Here, we group the proposed methods into two categories: those that assume the feedback given communicates {\em reward-like} information, and those that interpret the feedback as a {\em value-like} quantity.

\subsection{Human Feedback as Reward}
In the RL setting (a fixed MDP), a stationary reward function $R: \mathcal{S}\times\mathcal{A} \rightarrow \mathbb{R}$ is defined as a means by which to specify a fixed task.
Due to this stationarity, RL algorithms are able to seek policies that exhibit a notion of optimality with respect to a statistic dependent upon this distribution.
That is, RL algorithms seek $\pi^* = \text{arg} \; \text{max}_\pi J(\pi)$, where $J(\pi) = \mathbb{E}_\pi \left[ \sum_t \gamma^t R(s_t,a_t) \right] $ is well-defined.
The RL community has proposed several algorithms to accomplish this task, including policy gradient techniques \citep{sutton2000policy} and actor-critic techniques \citep{grondman2012survey}.

Due to the success of RL, some researchers have proposed techniques for learning from human feedback that interpret the feedback as the reward function itself \citep{isbell2001social,tenorio2010dynamic}.
Intuitively, this interpretation amounts to assuming that the human feedback provides an instantaneous rating of the agent's current decision.
For example, \cite{pilarski2011online} propose a technique for learning from human feedback that uses the feedback as the reward function in an actor-critic algorithm. During training, the human trainer provides a positive ($r=+0.5$) or negative ($r=-0.5$) reward to the learning agent. In the absence of a human-delivered reward, they simply assume the feedback is neural ($r=0$). For their experimental setting, they find that the proposed algorithm can learn useful policies when the human feedback is consistent, but that policy quality degrades when the feedback becomes inconsistent (i.e., becomes less stationary).
Another such technique is Advise \citep{griffith2013policy,cederborg2015policy}, in which the human feedback is used as the reward signal in a policy-gradient-like algorithm.
More specifically, the probability of an action being a good action is: 
\begin{equation}
 P_c(a) = \frac{\mathcal{C}^{\Delta s,a}}{\mathcal{C}^{\Delta s,a} + (1-\mathcal{C})^{\Delta s,a}}   
\end{equation}
where $\Delta s,a$ is the difference between the number of ``right'' and ``wrong'' labels that the human provided for action $a$ in state $s$. Notably, this approach coarsely takes human error into consideration using the $\mathcal{C}$ parameter. Let $\mathcal{C}$ denote the probability that an evaluation of an action choice is correctly provided by the human teacher ($\mathcal{C}$ = 0.5 is a random non-informative teacher, and $\mathcal{C}$ = 1 is a flawless teacher)~\citep{cederborg2015policy}. Suppose $P_q(a)$ is the probability of selecting action $a$ by an RL agent (e.g., according to Boltzmann distribution, Eq.~\ref{eqn:softmax}), the final policy during learning is determined using both $P_c$ and $P_q$:
\begin{equation}
\pi(s,a) = \frac{P_q(a)P_c(a)}{\sum_{a' \in \A} P_q(a')P_c(a') }
\end{equation}
In this way the algorithm combines knowledge it learned from interacting with the environment and knowledge it gained from human evaluative feedback.

\subsection{Human Feedback as Value}
Alternatively, several methods interpret the feedback signal as a value-like quantity~\citep{knox2009interactively,macglashan2017interactive}. Intuitively, this interpretation amounts to assuming that the human feedback provides a rating of the agent's current decision with respect to some forecast of future behavior.

One such technique is the TAMER algorithm (training an agent manually via evaluative reinforcement) \citep{knox2009interactively}, in which it is assumed that the human has in mind a desired policy $\pi_H$, and the feedback given at a time instant $t$, $H(s_t,a_t)$ roughly corresponds to $Q^{\pi_H}(s_t,a_t)$ (defined in Eq.~\ref{eq:q_value}).
TAMER agents use supervised learning with all the feedback collected up to time $t$ to calculate the current estimate of $H$, $\hat H$, e.g., through minimizing a standard squared loss~\citep{warnell2018deep}:
\begin{equation}
     \hat H^* = \arg\min_{\hat H} \sum_t \Big[\hat H(s_t,a_t) - H(s_t,a_t)\Big]^2
\end{equation}
Then the agent acts, in the next state, according to the policy 
\begin{equation}
    a_{t+1} = \arg \max_a \hat H^*(s_{t+1},a)
\end{equation}
in a fashion similar to Q Learning since we interpret $\hat H^*$ as an approximation for $Q^{\pi_H}$. Because the TAMER algorithm interprets the human feedback to be the value corresponding to a fixed (ideal) policy $\pi_H$, the implicit assumption made is that the feedback given is independent of the agent's current policy and depends only on the quality of an agent’s action
selection. This type of human feedback model is called \emph{policy-independent} models.

Alternatively, we could have \emph{policy-dependent} models in which the feedback depends on the agent’s current policy. An action selection may be rewarded or punished more depending on how often the agent would typically be inclined to select it. For example, the human may greatly reward the agent for deviating from its current policy to take a slightly better action (though this action may still be sub-optimal), and stop rewarding this action as the agent consistently adopts this action~\citep{macglashan2017interactive}. This phenomenon is known as diminishing returns and is policy-dependent~\citep{macglashan2017interactive}. The COACH (convergent actor-critic by humans) framework has leveraged the idea of policy-dependent feedback and assumes instead that the human feedback corresponds to the {\em advantage} (Eq.~\ref{eq:a_value}) for the current policy~\citep{macglashan2017interactive}. Intuitively, the advantage function communicates how much better or worse the agent's behavior is when deviating from its current policy. Algorithmically, COACH uses the feedback to replace the advantage function in calculating the policy gradient in an advantage actor-critic algorithm. Note that the human trainers do not need to provide feedback at every timestep like other evaluative feedback approaches.

With the advent of deep learning, several researchers in the community have recently begun attempting to use these techniques in the context of more challenging, high-dimensional state spaces.
For example, \cite{warnell2018deep} propose a technique that enables the use of TAMER for pixel-level state spaces in Atari games. To overcome the difficulty faced by trying to learn functions over such state spaces from sparse feedback, the authors propose to use a combination of a pre-trained deep autoencoder for state representation and a feedback replay buffer to allow for off-policy updates. \cite{arumugam2019deep} report that a similar approach is successful when applying COACH to pixel-level state spaces.
Aside from demonstrating the utility of learning from human feedback algorithms in high-dimensional state spaces, \cite{warnell2018deep} also reported that agents trained using human-provided feedback ultimately learned policies that outperformed that of the human trainers themselves.
This result would seem to support the hypothesis that the performance of agents that can learn from human feedback is not capped by the trainer's expertise to perform the task. However, such performance could be affected by the trainer's expertise in providing evaluative feedback.

\subsection{Extensions and Outlook}
\label{sec:feedback_extension}
Several extensions to the above algorithms have been proposed in the literature.
Notably, several have studied combining human-provided evaluative feedback with existing reward functions \citep{cederborg2015policy,knox2010combining,knox2012reinforcement,arakawa2018dqn} with the goal of augmenting reinforcement learning.
\cite{saunders2018trial} look explicitly at situations in which humans block catastrophic actions, and interpret these blocking actions as evaluative feedback when learning in combination with an existing reward function. This method is particularly useful for RL tasks that require safe exploration.

Evaluative feedback is usually communicated through button presses by humans, some other works have sought to infer feedback signals from multi-modal evaluative signals
humans naturally emit during social interactions, including gestures~\citep{najar2020interactively}, facial expressions~\citep{broekens2007emotion,arakawa2018dqn,cui2020empathic}, electroencephalogram (EEG) based brain waves signals~\citep{xu2020playing,akinola2020accelerated}, and implied feedback when humans refrain from giving explicit feedback~\citep{loftin2014learning,joachims2017accurately}. Other body-language and vocalization modalities not aimed at explicit communication, such as tone of voice, subtle head gestures, and hand gestures, could also be modeled and leveraged during training in the future~\citep{cui2020empathic}. Other works have looked at methods by which to elicit more feedback from human trainers \citep{li2016using} or to explicitly account for situations in which the human trainer may not be paying attention \citep{KesslerFaulkner:2019:AAP:3306127.3331762}.

Each of the algorithms presented above interprets human feedback in slightly different ways, resulting in different policy update rules.
Using synthetic feedback, \cite{macglashan2017interactive} showed that the convergence of these algorithms depends critically on whether the actual feedback matches the assumed one.
Critically, the nature of the feedback could potentially vary across tasks and trainers. \cite{loftin2016learning} has shown that instead of providing balanced feedback, human trainers could be more reward-focused (provides explicit rewards for correct actions and ignore incorrect ones), punishment-focused, or even inactive. Factors such as previous experience in training pets and feedback from the agent can affect the trainer's strategies~\citep{loftin2016learning}. Additionally, the nature of the feedback can be altered by the instruction given to the trainers. For example, \cite{cederborg2015policy} has shown that they can manipulate the meaning of human trainer's silence by differing the instructions given: The trainers were told that their silence meant positive/negative to the agent. Not surprisingly, the agent needs to adjust their interpretations of silence accordingly to perform well~\citep{cederborg2015policy}. Therefore, these factors need to be carefully controlled in practice.
One potential future research direction is to study methods that explicitly attempt to be robust to many types of feedback or methods that attempt to infer the human feedback type and adapt to that type in real time~\citep{grizou2014interactive,loftin2016learning,najar2020interactively}.

\section{Learning from Human Preference}
The second type of human guidance we discuss is that communicated in the form of a preference. As with evaluative feedback, for many tasks that we may wish artificial agents to learn, it may be difficult or impossible for humans to provide demonstrations due to challenges such as embodiment mismatch.
For example, consider control tasks with many degrees of freedom in which the artificial agent exhibits non-human morphology, as are commonly present in the MuJoCo environment ~\citep{todorov2012mujoco}.
Further, because of the complexity of the state space, it may also prove difficult for humans to provide fine-grained evaluative feedback on any particular portion of the behavior.

\begin{figure}
    \centering
    \includegraphics[width=1\textwidth]{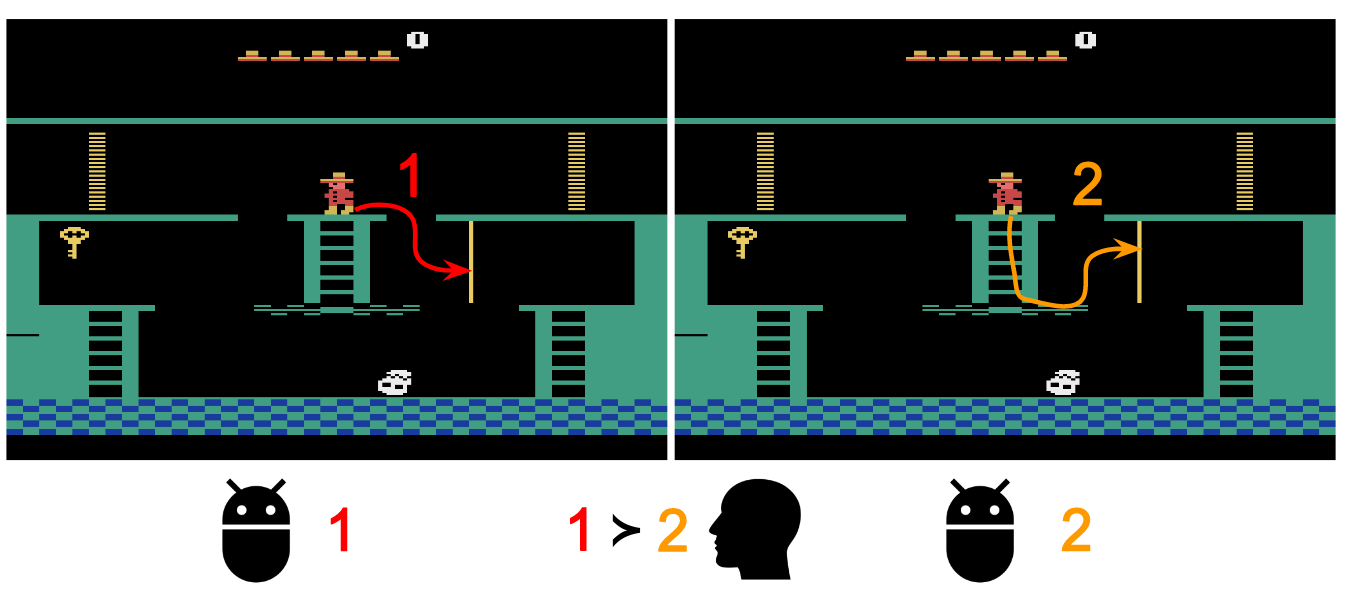}
    \caption{In learning from human preference, the learning agent presents two learned behavior trajectories to the human trainer, and the human tells the agent which trajectory is preferable. Here the human trainer prefers trajectory 1.}
    \label{fig:mtzm_pref}
\end{figure}

For such cases, some in the research community have posited that it is more natural for the agent to query human trainers for their \emph{preferences}, or \emph{rankings}, over a set of exhibited behaviors. This feedback can be provided for a set of state or action sequences; however, it is much less demanding if it is over trajectories as the trainer can directly evaluate outcomes.
Here, as shown in Fig.~\ref{fig:pref} and ~\ref{fig:mtzm_pref}, we consider preferences over trajectory segments, or sequences of state-action pairs: $\tau = ((s_0,a_0),(s_1,a_1),\dots)$. Using this type of human guidance, the learning problem is to learn a policy or an external reward function from human preference.

Preference learning has long been a topic of interest in the research community. Previous works have used preferences to directly learn policies \citep{wilson2012bayesian,busa2013preference}, learn a preference model \citep{furnkranz2012preference}, or learn a reward function \citep{wirth2016model,akrour2014programming}. A survey on these topics is provided by~\cite{wirth2017survey}.

More recent works have extended previous preference-based learning methods to be compatible with deep RL. The goal is to learn a hypothesized latent human reward function, $r(s,a)$ (as in IRL) from the communicated preferences~\citep{christiano2017deep,sadigh2017active,bestick2018learning,cui2018active}. In~\cite{christiano2017deep}, a pair of agent trajectories approximately 1-2 seconds in duration is simultaneously presented to human trainers to query for their preference. Under the model developed by~\cite{christiano2017deep}, the probability of a human preferring a segment depends exponentially on the total reward summed over the trajectory:
\begin{equation}
P \lbrack \tau^1 \succ \tau^2 \rbrack = \frac{\exp\sum r(s^1_t,a^1_t)}{\exp\sum r(s^1_t,a^1_t) + \exp\sum r(s^2_t,a^2_t)} 
\end{equation}
This model provides a training objective that can be used to find the reward function by minimizing the cross-entropy loss between the model's prediction and the human's preferences.
Since the targets to be evaluated are trajectories instead of state-action pairs, the feedback is typically very sparse compared to the amount of state-action data, resulting in a drastic reduction in human effort. The amount of human feedback required can be as little as 1\% of the total number of agent actions~\citep{christiano2017deep}. 

Preference learning problems are generally formulated as IRL problems. Hence it is natural to integrate preference and action demonstration via a joint IRL framework~\citep{palan2019learning,biyik2020learning}, with a nice insight that these two sources of information are complementary under the IRL framework: ``demonstrations provide a high-level initialization of the human's overall reward functions, while preferences explore specific,
fine-grained aspects of it"~\citep{biyik2020learning}. Therefore they use demonstrations to initialize a reward distribution, and refine the reward function with preference
queries~\citep{palan2019learning,biyik2020learning}. \cite{ibarz2018reward} takes a different approach to combine demonstration and preference information, by using human demonstrations to pre-train the agent. Further, they include demonstration trajectories when learning preferences, assuming human trajectories are always more preferable than agent trajectories. 

A key aspect of learning methods designed to leverage preferences is that of query selection, i.e., the decision the agent makes regarding which trajectories to query for the human's preference. \cite{christiano2017deep} select trajectories such that an ensemble of their learning models have the largest variance, i.e., uncertainty, in predicting the human's preference. Ideally, however, the query should maximize the expected information gain from an active learning perspective~\citep{cui2018active}, an important research challenge that is closely related to preference elicitation~\citep{zintgraf2018ordered}. \cite{sadigh2017active} have shown that query selection can be done by actively synthesizing preference queries. The reward learning procedure can be facilitated if selected queries can remove the maximal amount of hypotheses in the space of possible reward functions~\citep{sadigh2017active}. Follow-up works have extended this approach to batch-active methods~\citep{biyik2018batch}, using rankings instead of pairwise comparisons~\citep{biyik2019green}, and modeling the reward using more expressive models such as Gaussian processes~\citep{biyik2020active}. Query selection in preference learning falls into the general active learning paradigm and worth further investigation. 

\subsection{Extensions and Outlook}
In learning from human preferences, the targets to be evaluated are trajectories instead of state-action pairs as in learning from human evaluative feedback. The advantage of evaluating trajectories is that the feedback is typically very sparse, resulting in a drastic reduction in human effort. However, there are two potential concerns in leveraging human preference. First, although the amount of feedback is less, the time or cognitive efforts in watching the two trajectories to make a preference choice could be more. Second, selecting the optimal trajectory length is challenging. Shorter trajectories allow humans to provide feedback of high granularity at the cost of more frequent human interactions. From the human trainer's perspective, the ideal trajectory length should be \emph{subjective}, meaning that human trainers could select and adjust their preferred trajectory length during the training process. From the learning agent's perspective, the ideal length could also be \emph{adaptive}, meaning that the agent could adaptively adjust the trajectory length to query humans to receive feedback of desired granularity. 

Recent work by~\cite{bhatia2020preference} raised an important issue in preference learning: the human preference could be multi-criteria in nature, i.e., different behaviors are preferred under different criteria. For example, a driving policy $\tau_1$ is preferred in terms of comfort, while $\tau_2$ is preferred when speed is the only concern. Interestingly, a third policy $\tau_3$ which is a linear combination of $\tau_1$ and $\tau_2$ may be preferred among all three when considering both criteria~\citep{bhatia2020preference}. The authors propose a novel framework to solve this problem by decomposing the single overall comparison and ask humans to provide preferences along simpler criteria~\citep{bhatia2020preference}. The authors lay the groundwork from a game-theoretic perspective but many questions are yet to be answered.

\section{Hierarchical Imitation}
Many sequential decision-making tasks are hierarchically structured, meaning that they can be decomposed into subtasks and solved using a divide-and-conquer approach. As an example from behavioral psychology, case studies with non-human primates have shown that fine-grained, low-level actions are mostly learned without imitation. In contrast, coarse, high-level ``programs" learning is pervasive in imitation learning~\citep{byrne1998learning}. Program-level imitation is defined as imitating the high-level structural organization of a complex process, by observation of the behavior of another individual, while furnishing the exact details of actions by individual learning~\citep{byrne1998learning}, perhaps through reinforcement learning. 

\begin{figure}
    \centering
    \includegraphics[width=0.75\textwidth]{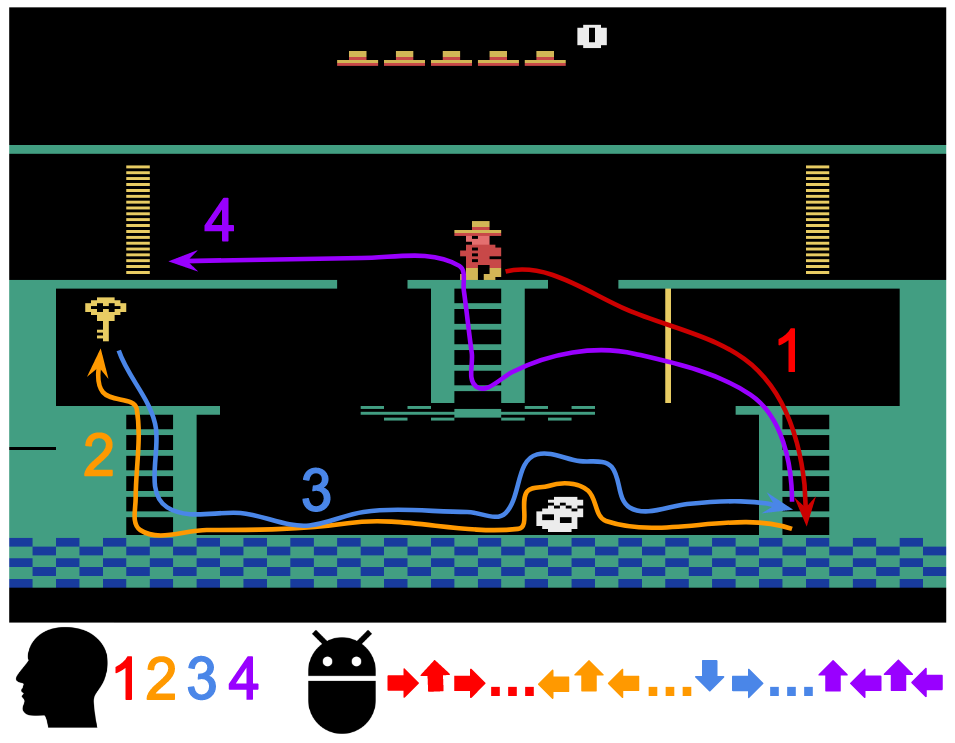}
    \caption{In hierarchical imitation, the basic idea is to have the human trainer specify high-level goals. For example, red goal one is to reach the bottom of the ladder. An agent will learn to accomplish each high-level goal by performing a sequence of low-level actions potentially through reinforcement learning by itself.}
    \label{fig:mtzm_hi}
\end{figure}

Therefore, an interesting form of guidance can be provided by asking human trainers to provide only high-level feedback on these tasks. Similar to preference, this type of feedback also targets trajectory segments but is provided as choices of high-level goal in a given state, such as options\footnote{An option is a temporally extended action, or macro-action, which is composed of a policy, a termination condition, and an initiation set~\citep{sutton1999between}.}. Due to the hierarchical structure of the task, the behavior trajectory can be naturally segmented into options, instead of arbitrary segments in the preference framework. As shown in Fig.~\ref{fig:hi} and ~\ref{fig:mtzm_hi}, using this type of human guidance, the learning problem for the agent is to learn a policy for choosing high-level goals in addition to learning a policy for low-level action selection.

\subsection{Simulated Agents}
\cite{le2018hierarchical} has proposed a hierarchical guidance framework that assumes a two-level hierarchy, in which a high-level agent learns to choose a goal $g$ given a state, while low-level agents learn to execute a sub-policy (option) to accomplish the chosen goal (Fig.~\ref{fig:hi}). Note that an action to terminate the current option needs to be added to the low-level agent's action space, and this termination action can be demonstrated by a human and learned by the agent. Human trainers were asked to provide three types of feedback: 1) a positive signal if the high-level goal $g_t$ and low-level sub-policy $a_t$ are both correct; 2) the correct high-level goal if the chosen one is incorrect; 3) the correct low-level action $a^*_t$ if the high-level goal was chosen correctly but the low-level sub-policy is incorrect. At each level, the learning task becomes a typical imitation learning problem, therefore conventional IL algorithms such as behavioral cloning and DAgger~\citep{ross2011reduction} can be applied. 

Perhaps the most exciting result comes from a hybrid of hierarchical imitation learning and RL. One approach is to train the agent to choose high-level goals via imitation learning, and let the agent learn low-level policies via RL by itself. This approach was shown to be substantially more sample efficient than conventional imitation learning. For example, \cite{andreas2017modular} only required humans to provide policy sketches which are high-level symbolic subtask labels. The policy of each subtask is learned by the RL agent on its own and no longer requires human demonstration. A similar approach has been shown to be successful on Atari Montezuma's Revenge~\citep{le2018hierarchical}. Another approach is to train both high-level and low-level policies via imitation learning, then fine-tune them using RL later~\citep{gupta2020relay}.

\subsection{Physical Robots} 
Compared to simulated learning agents, in physical robot learning safety and sample efficiency are critical issues for IL and RL algorithms. Therefore, incorporating human prior knowledge through hierarchical task structuring to make robot learning tractable has been long studied and implemented. The classic approach is to define and represent the learning task at a more abstract level using human knowledge. As an example, actions can be defined as high-level goals such as ``turn 90 degrees clockwise", meanwhile fine-grained motor commands that accomplish these goals can be handled by low-level controllers. An early survey of previous robotic research on this topic is provided by~\cite{kober2013reinforcement}. 

In contrast to providing only high-level goals for simulated agents~\citep{andreas2017modular}, in robotic tasks humans often need to provide low-level demonstrations due to safety and sample efficiency concerns. The works that leverage this type of demonstration can be roughly classified into \emph{segmentation-based} or \emph{non-segmentation-based} approaches depending on whether the task hierarchy is provided by humans. 

In segmentation-based approaches, the task hierarchy is not provided, hence the aim is to extract meaningful segments from the low-level demonstration trajectories. For example, in contrast to \cite{andreas2017modular}, \cite{krishnan2017ddco} and \cite{henderson2018optiongan} set up the learning task in the opposite way which attempts to discover high-level options from low-level demonstration data. In this setting, only low-level demonstrations are collected, the options are latent variables of the trainer that can be inferred in a fashion similar to Expectation Maximization~\citep{krishnan2017ddco}. Similarly, other methods aim to learn low-level primitives~\citep{kipf2019compile,sharma2018directed}, latent conditioned policies~\citep{hausman2017multi}, goal-conditioned policies~\citep{gupta2020relay}, or skills~\citep{konidaris2012robot,kroemer2015towards} which meaningfully segment the low-level demonstrations. If information about high-level goals is also provided, they can be used to help infer meaningful segmentation boundaries. For example, \cite{codevilla2018end} has successfully combined high-level navigation commands with low-level control signals in a framework named conditional imitation learning for autonomous driving tasks. 

In contrast, the task hierarchy can be provided explicitly or implicitly to eliminate the need for task segmentation. Humans can explicitly define the task hierarchy and feed such information to the robots~\citep{mohseni2015interactive}. Alternatively, humans can provide demonstrations subtask by subtask, therefore options, subtasks, subprograms, subroutines, or individual skills are learned first in isolation and then combined~\citep{friesen2010imitation}. 

A fixed, rigid task hierarchy has a poor ability to generalize. Recently, a framework named neural programming (NTP) has been developed that can decompose a demonstrated task into modular and reusable neural programs in a hierarchical manner~\citep{reed2015neural,li2016neural,xu2018neural,fox2018parametrized}. The task hierarchy is only provided by humans during the training phase until the agent has learned to do task segmentation on its own. Neural programs are structured policies that perform algorithmic tasks by controlling the behavior of a computation mechanism~\citep{fox2018parametrized}. They are represented by neural networks that can learn to represent and execute compositional programs from demonstrations~\citep{reed2015neural}. The demonstration is still provided as low-level actions, the learning algorithm attempts to learn and reuse primitive network modules from the demonstration.  A task manager, which is often a trainable task-agnostic network, decides which subprogram to run next and feeds task specification to the next program. Training this high-level manager requires ground-truth high-level task labels provided by human~\citep{xu2018neural}. The low-level policy is represented as a neural program that takes a task specification as its input argument. 

The benefit of this kind of hierarchical modular approach comes from the observation that in many robotic tasks there are shared components, and a learned task component (e.g., a particular skill) can often generalize across tasks. In a multitask setting, learned task components can be transferred between tasks so the required human demonstration effort could be drastically reduced~\citep{fox2019multi,xu2018shared}.


\subsection{Extensions and Outlook}
In some of the above robot learning works, asking humans to provide high-level actions requires additional human effort. However, in physical robot experiments, human annotation is often less costly and risky than demonstrations or teleoperations. If providing extra annotation can reduce the cost of using physical robots, such extra effort is justified and desirable~\citep{fox2019multi}.

We have seen that humans can either provide low-level action demonstrations, or high-level goals, or both types of guidance together. The choice that is suitable for a particular task domain depends on at least two factors. The first concern is the relative effort in specifying goals vs. providing demonstrations. High-level goals are often clear and easy to be specified in tasks such as navigation~\citep{andreas2017modular}. On the contrary in tasks like Tetris providing low-level demonstration is easier since high-level goals are not easy to represent and be communicated. The second concern is safety and sample efficiency. Only providing high-level goals requires the agents to learn low-level policies by themselves through trial-and-error, perhaps with many more samples, which is suitable for simulated agents but not for physical robots. Therefore in robotic tasks, low-level action demonstrations are often required.

One way to further reduce human effort in hierarchical imitation learning is to leverage evaluative feedback. Evaluative feedback can be naturally incorporated in the hierarchical imitation learning framework, in which human trainers provide evaluative feedback (yes or no) on either high-level or low-level actions of the learning agent, an approach that has been partially explored by~\cite{mohseni2015interactive} and \citep{le2018hierarchical}. Moreover, as mentioned earlier, it is natural to extend hierarchical imitation to incorporate human preferences over the outcome of options, instead of asking humans to provide the correct option labels, as done in~\cite{pinsler2018sample}. 

Hierarchical imitation learning is naturally related to hierarchical reinforcement learning~\citep{sutton1999between,dietterich2000hierarchical,barto2003recent}, which is an active research field with its own exciting progress~\citep{kulkarni2016hierarchical,bacon2017option,vezhnevets2017feudal,nachum2018data}. Another closely related research field here is multi-agent reinforcement learning~\citep{ghavamzadeh2006hierarchical} since the multi-agent setting implicitly contains a two-level hierarchy: One at the individual agent's level and the other at the group level. For a recent survey on this topic, please see~\cite{hernandez2019survey}. A potential research direction is to leverage human guidance, in the forms of demonstration or feedback, in the settings of hierarchical RL or multi-agent learning systems. As we have seen in~\cite{le2018hierarchical}, a reasonable starting point is to ask humans to demonstrate or evaluate high-level decisions.

\section{Imitation from Observation}
\begin{figure}
    \centering
    \includegraphics[width=1\textwidth]{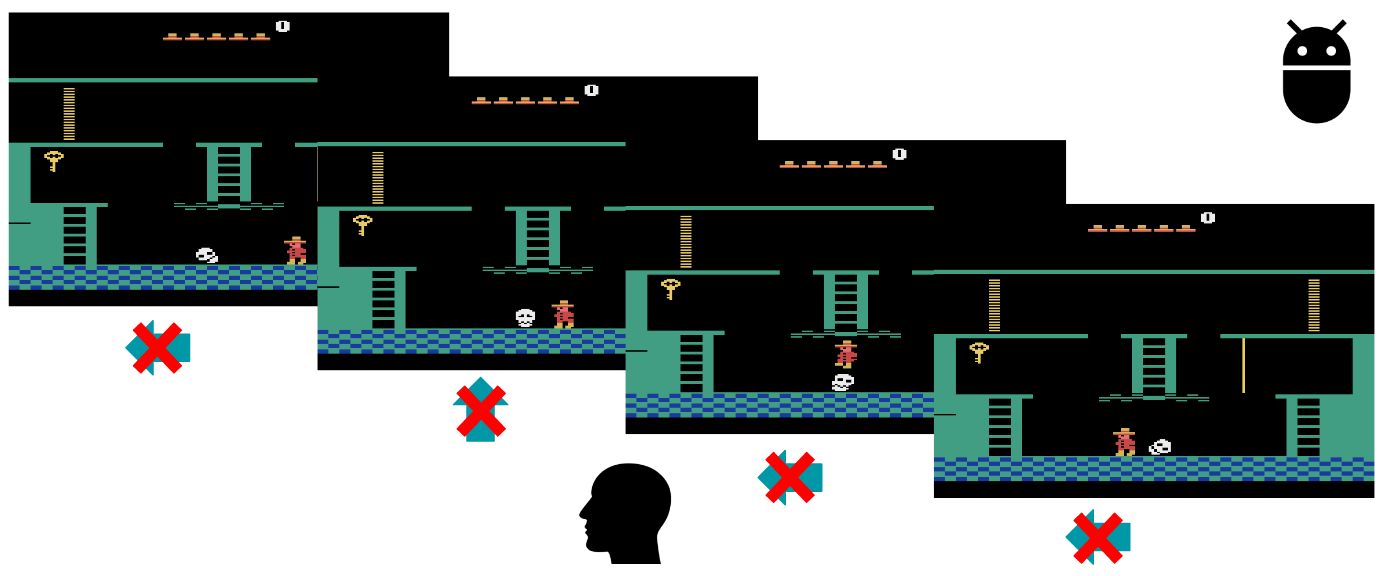}
    \caption{In imitation from observation, the setting is very much like standard imitation learning (Fig.~\ref{fig:mtzm_il}), except the agent does not have access to the actions demonstrated by the human trainer.}
    \label{fig:mtzm_ifo}
\end{figure}

Imitation from observation (IfO) \citep{torabi2019recent} is the problem of learning directly by observing a trainer performing the task. The learning agent only has access to state demonstrations (e.g. in the form of visual observations) of the trainer (Fig.~\ref{fig:ifo} and ~\ref{fig:mtzm_ifo}). Using this type of human guidance, the learning problem for the agent is to learn a policy from the state sequences demonstrated by the human. 

This framework is different from conventional imitation learning in the sense that it eschews the requirement for action labels in demonstrations. Removing this constraint enables imitating agents to use a large amount of previously ignored available demonstration data such as videos on YouTube. The ultimate goal in this framework is to enable agents to utilize the existing, rich amount of demonstration data that do not have action labels, such as the human guidance provided through online videos of humans performing various tasks.

Broadly speaking, there are two major components of the \IfO ~problem: {\em (1)} perception, and {\em (2)} control. 

\subsection{Perception}
Because \IfO~depends on observations of an expert agent, processing these observations perceptually is extremely important. Previous works have used multiple approaches for this part of the problem.
One such approach is to record the expert's movements using sensors placed directly on the expert agent \citep{ijspeert2001trajectory}.
Using these recordings, techniques have been proposed that can allow humanoid or anthropomorphic robots to mimic human motions, e.g., arm-reaching movements \citep{ijspeert2002movement,bentivegna2002humanoid}, biped locomotion \citep{nakanishi2004learning}, and human gestures \citep{calinon2007incremental}. 
Another approach to the perception problem is that of motion capture \citep{field2009motion}, which typically uses visual markers on the demonstrator to infer movement.
\IfO ~techniques built upon this approach have been used for a variety of tasks, including locomotion, acrobatics, and martial arts \citep{peng2018deepmimic,merel2017learning,setapen2010marionet}.
The methods discussed above often require costly instrumentation and pre-processing \citep{holden2016deep}, and therefore cannot be used in conjunction with more passive resources such as YouTube videos.

Recently, however, convolutional neural networks and advances in visual recognition have provided promising tools to work towards visual imitation where the expert demonstration consists of raw video information (e.g., pixel color values) alone.
Even with such tools, the imitating agent is still faced with several challenges: {\em (1)} embodiment mismatch, and {\em (2)} viewpoint difference.
Embodiment mismatch arises when the demonstrating agent has a different embodiment from that of the imitator.
For example, the video could be of a human performing a task, but the goal may be to train a robot to do the same.
Since humans and robots do not look exactly alike (and may look quite different), the challenge is in how to interpret the visual information such that \IfO ~can be successful.
One \IfO ~method developed to address this problem learns a correspondence between the embodiments using autoencoders in a supervised fashion \citep{gupta2017learning}.
The autoencoder is trained in such a way that the encoded representations are invariant with respect to the embodiment features.
Another method learns the correspondence in an unsupervised fashion with a small amount of human supervision \citep{sermanet2018time}.
The second \IfO~perceptual challenge is the viewpoint difference that arises when demonstrations are not recorded in a controlled environment.
For instance, the video background may be cluttered, or there may be a mismatch in the point of view present in the demonstration video and that with which the agent sees itself.
One \IfO ~approach that attempts to address this issue learns a context translation model to translate an observation by predicting it in the target context \citep{liu2018imitation}.
The translation is learned using data that consists of images of the target context and the source context, and the task is to translate the frame from the source context to that of the target.
Another approach uses a classifier to distinguish between the data that comes from different viewpoints and attempts to maximize the domain confusion in an adversarial setting during the training \citep{stadie2017third}.
Consequently, the extracted features can be invariant with respect to the viewpoint.


\begin{figure*}[!ht]
	\centering
	\begin{tikzpicture}[thick,scale=.6, every node/.style={scale=.6}]

	\draw[use as bounding box, transparent] (-3,1.5) rectangle (16.6, 4);
	\draw[rounded corners,fill=light-blue,draw=black] (3.4,3.5) -- (3.4,4.1) -- (8.6,4.1) -- (8.6,3.5) -- cycle;
	\node [align=center] at (6, 3.75) {\LARGE \IfO ~Control Algorithms};
	
	\draw[->, line width=0.2mm] (6,3.5) -- (1,3);
	\draw[rounded corners,fill=light-blue,draw=black] (-.7,2.5) -- (-.7,3) -- (2.7,3) -- (2.7,2.5) -- cycle;
	\node [align=center] at (1, 2.75) {\LARGE Model-based};
	
	\draw[->, line width=0.2mm] (6,3.5) -- (11,3);
	\draw[rounded corners,fill=light-blue,draw=black] (9.5,2.5) -- (9.5,3) -- (12.5,3) -- (12.5,2.5) -- cycle;
	\node [align=center] at (11, 2.75) {\LARGE Model-free};
	
	\draw[->, line width=0.2mm] (1,2.5) -- (-1,2);
	\draw[rounded corners,fill=light-blue,draw=black] (-2.7,1.5) -- (-2.7,2) -- (.7,2) -- (.7,1.5) -- cycle;
	\node [align=center] at (-1, 1.75) {\LARGE Inverse Model};
	
	\draw[->, line width=0.2mm] (1,2.5) -- (3,2);
	\draw[rounded corners,fill=light-blue,draw=black] (1.2,1.5) -- (1.2,2) -- (4.8,2) -- (4.8,1.5) -- cycle;
	\node [align=center] at (3, 1.73) {\LARGE Forward Model};
	
	\draw[->, line width=0.2mm] (11,2.5) -- (8,2);
	\draw[rounded corners,fill=light-blue,draw=black] (5.65,1.5) -- (5.65,2) -- (10.35,2) -- (10.35,1.5) -- cycle;
	\node [align=center] at (8, 1.75) {\LARGE Adversarial Methods};
	
	\draw[->, line width=0.2mm] (11,2.5) -- (14,2);
	\draw[rounded corners,fill=light-blue,draw=black] (11.75,1.5) -- (11.75,2) -- (16.25,2) -- (16.25,1.5) -- cycle;
	\node [align=center] at (14, 1.73) {\LARGE Reward-Engineering};

	\end{tikzpicture}
	\caption{A diagrammatic representation of categorization of the \IfO ~control algorithm. The algorithms can be categorized into two groups: (1) model-based algorithms in which the algorithms may use either a forward dynamics model \protect\citep{edwards2018imitating} or an inverse dynamics model \protect\citep{torabi2018behavioral,nair2017combining}. (2) Model-free algorithms, which itself can be categorized into adversarial methods \protect\citep{torabi2018generative,merel2017learning,stadie2017third} and reward engineering \protect\citep{sermanet2018time,gupta2017learning,liu2018imitation}.}
	\label{fig:alg}
\end{figure*}
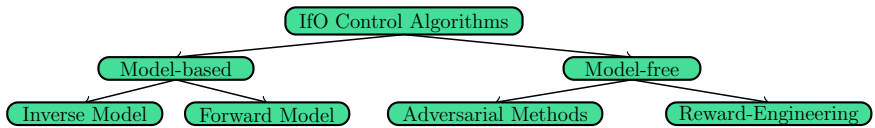

\subsection{Control}
Another main component of \IfO ~is control, i.e., the approach used to learn the imitation policy, typically under the assumption that the agent has access to clean state demonstration data $\{s_t\}$.
Since the action labels are not available, this is a very challenging problem, and many approaches have been discussed in the literature.
Previously, this problem was referred to as {\em trajectory tracking} where the goal was to follow a time parameterized reference \citep{yang1999sliding,aguiar2007trajectory}. Most of the algorithms developed for trajectory tracking do not involve machine learning at all and require the state features and reference points to be well-defined such as joint angles, velocities, etc. \citep{yang1999sliding,caracciolo1999trajectory}. Therefore, it is not clear how to directly scale these algorithms to visual imitation (imitating directly from raw pixel data).

With the rise of deep learning, however, many new learning-based algorithms have recently been proposed to tackle the \IfO ~control problem. We organize them here into two general groups: {\em (1)} model-based algorithms, and {\em (2)} model-free algorithms.
Model-based approaches to \IfO ~are characterized by the fact that they learn some type of dynamics model during the imitation process. Most of these algorithms learn an inverse dynamics model which is a mapping from state-transitions $\{(s_t, s_{t+1})\}$ to actions $\{a_t\}$ \citep{hanna2017grounded}. The goal of these algorithms is to retrieve the missing demonstration action labels. To do so, they interact with the environment, collect state action data, and then learn an inverse dynamics model. Applying this learned model on two consecutive demonstrated states would output the missing taken action that had resulted in that state transition. After retrieving the actions, the learning problem can be treated as a conventional imitation learning problem. Recently, many algorithms are developed with this high-level idea \citep{nair2017combining,torabi2018behavioral,pavse2020ridm,pathak2018zeroshot,guo2019hybrid,robertson2020concurrent,jiang2020offline,radosavovic2020state}. Some other algorithms use forward dynamics model instead which is a mapping from state-action pairs, $\{(s_t, a_t)\}$, to the next states, $\{s_{t+1}\}$. One algorithm of this type is developed by \citeauthor{wu2020model} [\citeyear{wu2020model}] in which a forward dynamics model is learned which is used to predict the future state of the agent and then future state similarity is used to learn an imitation policy.
There is another algorithm \citep{edwards2018imitating} that learns \emph{forward} dynamics model. This algorithm hypothesizes that the state transitions are caused by the actions taken by the agent. The actions are unknown and therefore the algorithm considers a latent (unreal) action space and learns a policy in that latent space that best describes the state transitions. Since the actions generated by this learned policy are not real, next the agent takes a few interactions with the environment to make corrections to the action labels. To be more specific, this algorithm creates an initial hypothesis for the imitation policy by learning a latent policy $\pi(z|s_t)$ that estimates the probability of latent (unreal) action $z$ given the current state $s_t$.
Since actual actions are not needed, this process can be done offline without any interaction with the environment.
To learn the latent policy, they use a latent forward dynamics model which predicts $s_{t+1}$ and a prior over $z$ given $s_t$.
Then they use a limited number of environment interactions to learn an action-remapping network that associates the latent actions with their corresponding correct actions.
Since most of the process happens offline, the algorithm is efficient with regard to the number of interactions needed.

The other broad category of \IfO ~control approaches is that of model-free algorithms.
Model-free techniques attempt to learn the imitation policy without any sort of model-learning step.
Within this category, there are two fundamentally different types of algorithms. One is adversarial methods which are inspired by the generative adversarial imitation learning (\GAIL) algorithm described in Section \ref{sec:IL}. In \GAIL, the goal is to bring the state-action distribution of the imitator close to that of the demonstrator. However, since in \IfO~the imitator does not have access to the actions, the proposed algorithms attempt to bring the state distribution \citep{merel2017learning,henderson2018optiongan}, or state transition distribution \citep{stadie2017third,torabi2018generative,torabi2019adversarial,torabi2019imitation,
torabi2019sample,zolna2018reinforced,sun2018provably,yang2019imitation,chaudhury2019injective} of the imitator close to that of the demonstrator. The overall scheme of these algorithms is as follows. They use a \GAN -like architecture in which the imitation policy is interpreted as the generator.
The imitation policy is executed in the environment to collect data, $\{(s_t^i, a_t^i)\}$, and either the states or the state transitions are fed into the discriminator, which is trained to differentiate between the data that comes from the imitator and data that comes from the demonstrator.
The output value of the discriminator is then used as a reward to update the imitation policy using RL. Another class of model-free approaches developed for \IfO ~control is that utilizes reward engineering.
Here, reward engineering means that, based on the expert demonstrations, a manually designed reward function is used to find imitation policies via \RL.
Importantly, the designed reward functions are not necessarily the ones that the demonstrator used to produce the demonstrations---rather, they are simply estimates inferred from the demonstration data. Most of the algorithms of this type use the negative of the Euclidean distance of the states of the imitator and the demonstrator  (or an embedded version of them) as the reward at each time step \citep{kimura2018internal,sermanet2018time,dwibedi2018learning,gupta2017learning,liu2018imitation,liu2020hilonet}. Another approach of this type is developed by \cite{goo2018learning} in which the algorithm uses a formulation similar to shuffle-and-learn \cite{misra2016shuffle} to train a neural network that learns the order of frames in the demonstration. The network in a supervised fashion gets two observations and outputs a value between zero and one. The closer the value to one, the higher the chance of observations being in the right order. This neural network is then used as a surrogate reward function to train a policy.
\cite{aytar2018playing} also take a similar approach, learning an embedding function for the video frames based on the demonstration. 
They use the closeness between the imitator's embedded states and some checkpoint embedded features as the reward function.

\subsection{Extensions and Outlook}
Regarding the perception component of the \IfO~problem, adversarial training techniques have led to several recent and exciting advances in the computer vision community.
One such advance is in the area of pose estimation \citep{cao2017realtime,wang20193d}, which enables detection of the position and orientation of the objects in a cluttered video through keypoint detection---such keypoint information may also prove useful in \IfO.
While there has been a small amount of effort to incorporate these advances in \IfO ~\citep{peng2018sfv}, there is still much to investigate.

Another recent advancement in computer vision is in the area of visual domain adaptation \citep{wang2018deep}, which is concerned with transferring learned knowledge to different visual contexts.
For instance, the recent success of CycleGAN \citep{zhu2017unpaired} suggests that modified adversarial techniques may be applicable to \IfO ~problems that require solutions to embodiment mismatch, though it remains to be seen if such approaches will truly lead to advances in \IfO.

Regarding the control component of the \IfO~problem, very few of the mentioned \IfO ~algorithms discussed have been successfully tested on physical robots, such as \cite{sermanet2018time,liu2018imitation}.
That is, most discuss results only in simulated domains.
For instance, while adversarial control methods currently provide state-of-the-art performance for several baseline experimental \IfO ~problems, these methods exhibit high sample complexity and have therefore only been applied to relatively simple simulation tasks.
Thus, an open problem in \IfO ~is that of finding ways to adapt these techniques such that they can be used in scenarios for which high sample complexity is prohibitive, i.e., tasks in robotics. Furthermore, there have been few works investigating the combination of \IfO~with other learning frameworks \citep{brown2019extrapolating,pavse2020ridm,schmeckpeper2020reinforcement}. There is room to investigate how different types of learning paradigms could be incorporated in \IfO~to improve the overall task learning performance.

\section{Learning Attention from Humans}
During the human demonstration or evaluation process, there are other useful learning signals. One useful signal is human visual attention, which can be treated as a form of guidance that reveals important task features to the learning agent. For decision tasks with high-dimensional visual information as input, humans visual attention is revealed by eye movements, i.e., gaze behaviors. Gaze is an informative source of {\em (1)} important state features in high-dimensional state space at a given time {\em (2)} the explanatory information that reveals the target or goal of an observed action. For the former, since human eyes have limited resolution except for the center fovea, humans learn to move their eyes to the correct place at the right time to process urgent state information. For the latter, knowing which visual object the human trainer looked at while making a decision can help to explain why a particular decision was made. For these reasons, learning attention from humans could help a learning agent extract useful features from a high-dimensional state space and understand the underlying causes of a human trainer's demonstrated action. This approach has recently become very popular as learning agents migrate from simple tasks to challenging sequential decision-making tasks with high-dimensional inputs (Fig.~\ref{fig:tasks}).

\begin{figure}
    \centering
    \includegraphics[width=1\textwidth]{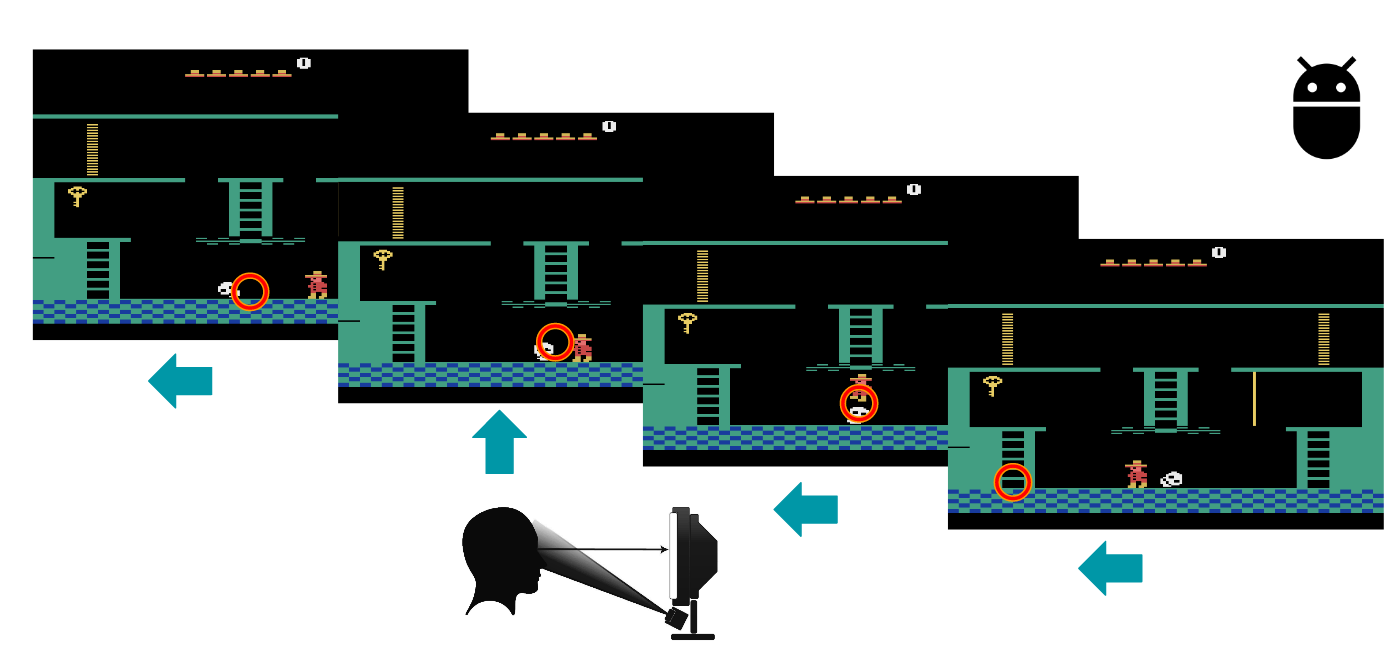}
    \caption{In learning attention from humans, the agent has access to human attention information in addition to the action demonstrations. The eye movement data indicated by the red circles here can be recorded by an eye tracker. This data reveals the current behavioral goal (such as the object of interest, e.g., the skull and the ladder) when taking an action.}
    \label{fig:mtzm_att}
\end{figure}

The gaze data can be collected with an eye tracker while the human trainer is demonstrating the task (Fig.~\ref{fig:att} and ~\ref{fig:mtzm_att}). Recently, researchers have collected human gaze and policy data for meal preparation \citep{li2018eye}, Atari game playing \citep{zhang2020atari}, human-to-human (non-verbal) interactions \citep{zuo2018gaze}, and outdoor driving \citep{palazzi2018predicting}. Using this type of human guidance, the learning problem for the agent is to learn the attention mechanism from humans in addition to learning a decision policy.

\subsection{Attention Learning}
The first learning objective using these datasets could be training an agent to imitate human gaze behaviors, i.e., learning to attend to certain features of a given image. The problem was formalized as a visual saliency prediction problem in computer vision research~\citep{itti1998model}. Recently this area has made tremendous progress due to deep learning as large-scale eye-tracking datasets became available for images~\citep{papadopoulos2014training,li2014secrets,xu2014predicting,bylinskii2015saliency,bylinskii2015intrinsic,krafka2016eye}, videos~\citep{mathe2014actions,wang2018revisiting}, and 360-degree videos~\citep{zhang2018saliency,xu2018gaze}. Visual saliency is a well-developed field in computer vision. We direct interested readers to recent review papers on the topics of saliency evaluation metrics~\citep{bylinskii2019different}, saliency model performance analyses~\citep{bylinskii2016should,he2019understanding}
and a closely related field called salient object detection~\citep{borji2015salient}.

\begin{figure*}[ht!]
\centering
\subfloat[Atari Ms.Pacman]{\includegraphics[width=0.495\textwidth]{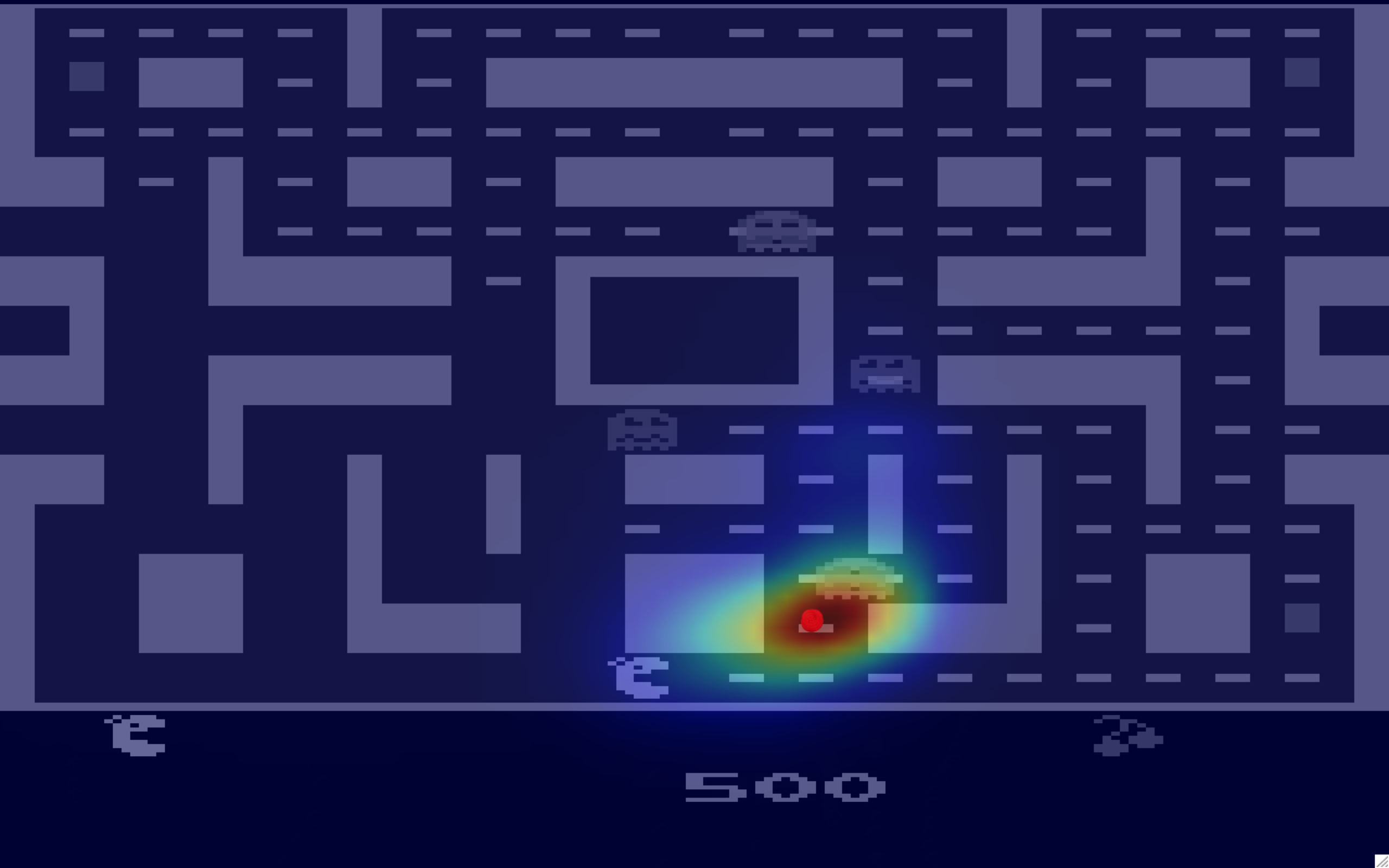}}
\hspace{0.001cm}
\subfloat[Atari Seaquest]{\includegraphics[width=0.495\textwidth]{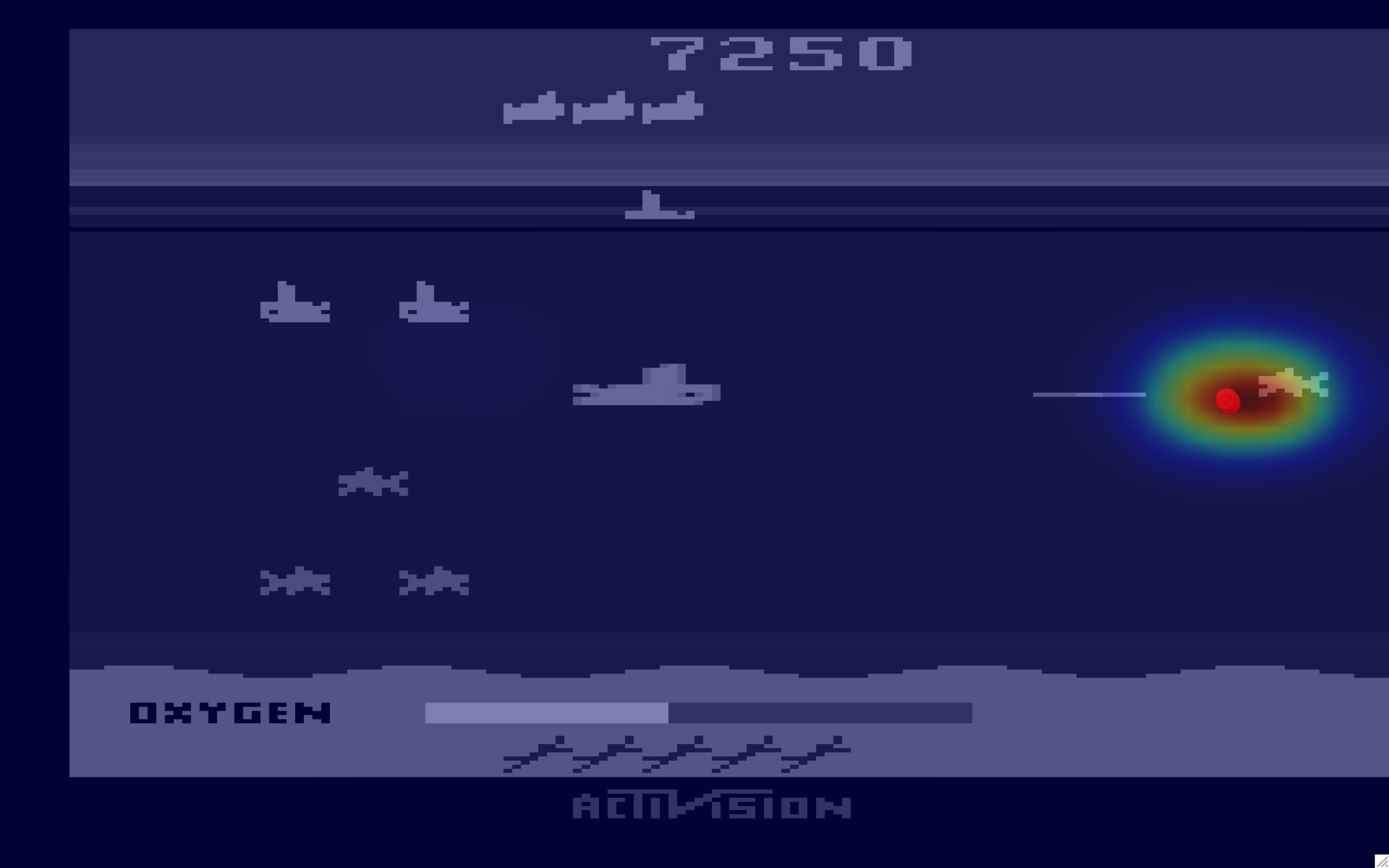}}\\
\subfloat[Meal preparation]{\includegraphics[width=0.75\textwidth]{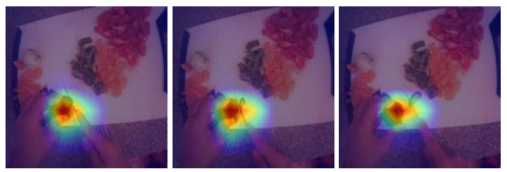}}\\
\subfloat[Driving]{\includegraphics[width=0.5\textwidth]{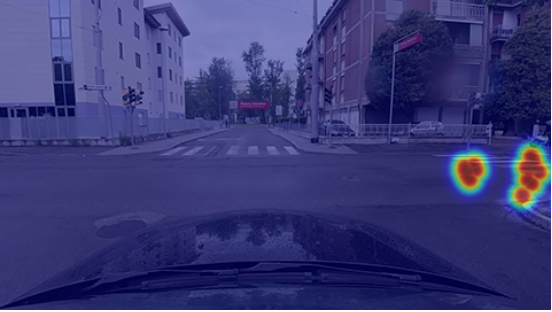}}
\subfloat[Driving]{\includegraphics[width=0.5\textwidth]{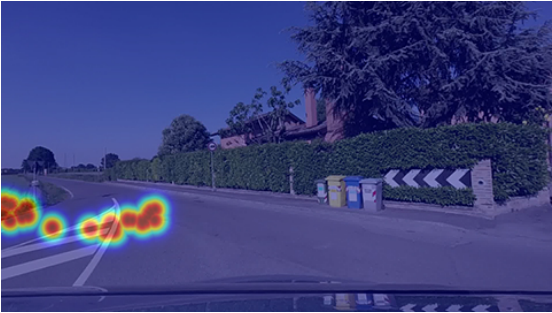}}
\caption{Learning attention from human in game playing~\citep{zhang2018agil}, meal preparation~\citep{li2018eye}, and driving~\citep{palazzi2018predicting}. The heatmaps show the agent's prediction of human attention, represented as saliency maps (probability distribution of attention) overlayed on the images. Red indicates regions that have high predicted probability to be attended by humans.}
\label{fig:gaze}
\end{figure*}

In our context of sequential decision-making tasks, the saliency prediction problem can be formalized as follows:
\begin{quote}
    Given a state $s_t$, learn to predict human gaze positions $w_t$, i.e., learn $P(w|s)$.
\end{quote}
Note that $w_t$ could be a set of positions since the human can look at multiple regions of the image. In practice, discrete human gaze positions are converted into a continuous distribution \citep{bylinskii2019different}. So the agent should learn to predict this probability distribution over the given image. This can be done using supervised learning where Kullback-–Leibler divergence can be used as the loss function to calculate the difference between the ground truth distribution $P$ and predicted distribution $Q$~\citep{bylinskii2019different}:
\begin{equation}
 KL(P,Q) = \sum_i \sum_j Q(i,j) \log \Big(\epsilon + \frac{Q(i,j)}{\epsilon + P(i,j)}\Big)
\end{equation}
where $i,j$ are pixels indices and $\epsilon$ is a small regularization constant and determines how much zero-valued predictions are penalized. Recent works have trained convolutional neural networks to accomplish this learning task \citep{li2018eye,zhang2020atari,palazzi2018predicting,deng2019drivers,chen2020robot}. Example gaze prediction results in the format of saliency maps can be seen in Fig.~\ref{fig:gaze}. A notable challenge here is \emph{egocentric} gaze prediction in which the spatial distribution of the gaze is highly biased towards the image center, a problem further addressed by~\cite{palazzi2018predicting,tavakoli2019digging}.

\subsection{Decision Learning}
In computer vision, traditional saliency prediction does not involve active tasks nor human decisions. The humans look at static images or videos in a free-viewing manner without performing any particular task and only the eye movements are recorded and modeled. Meanwhile, the aforementioned datasets all require humans to perform a task while collecting their gaze and action data. From a decision-learning perspective, human attention may provide additional information about their decisions, therefore it is intuitive to leverage learned attention models to guide the learning process of human decisions. The learning problem can be formalized as follows: 
\begin{quote}
    Given a state $s_t$ and human gaze positions $w_t$, learn to predict human action $a_t$, i.e., learn $P(a|s,w)$.
\end{quote}
Intuitively, knowing where humans would look provides useful information on what action they will take. To incorporate human attention into action learning, there are at least three common methods: as an additional channel of information, as a mask on the input to filter out unimportant information, or as a secondary optimization objective. For example, in training a neural network, the above methods correspond to concatenating a gaze map with the input image, masking the input image with the gaze map, and adding gaze prediction as an auxiliary loss term in the objective function, respectively~\citep{zhang2020human}. The most popular way is to treat the predicted gaze distribution of an image as a filter or a mask. This mask can be applied to the image to generate a representation of the image that highlights the attended visual features.

Experimental results have shown that including gaze information leads to higher accuracy in recognizing or predicting human actions, in reaching~\citep{ravichandar2018gaze}, human-to-human interaction~\citep{zuo2018gaze}, driving~\citep{xia2018predicting,liu2019gaze,chen2019gaze,xia2020periphery}, meal preparation~\citep{li2018eye,shen2018egocentric,sudhakaran2019lsta,huang2020mutual}, and video game playing~\citep{zhang2018agil,zhang2020atari}. 

Once the agent has learned both the attention and decision models from human data, it can perform the task on its own. It has been shown that incorporating a learned gaze model into imitation learning agents leads to a large performance increase, comparing to agents without attention information~\citep{zhang2020atari,saran2020efficiently,chen2020robot}. For real-world tasks like autonomous driving, it is reasonable to expect a similar improvement when incorporating human attention models. Due to physical constraints and safety reasons, this is yet to be explored but preliminary tests in simulated environments are possible.

\subsection{Extensions and Outlook}
In general, human gaze is a good indicator of the underlying decision-making mechanism, it bridges perception and decision-making by indicating the current behavioral target. The gaze data can be collected in parallel with actions. One concern with this approach is the hardware and software required to collect human gaze data. Recent progress in computer vision has improved eye tracker accuracy and portability by a significant margin. Appearance-based algorithms using convolutional neural networks have been shown to have better tracking accuracy and are more robust to visual appearance variations~\citep{zhang2015appearance,wood2015rendering,krafka2016eye,shrivastava2017learning,zhang2017mpiigaze,park2018deep}, compared to more traditional approaches like hand-crafted feature-based or model-based algorithms. Advanced tracking software can estimate gaze in real-time from head poses and appearance without specialized hardware on low-cost devices such as webcams~\citep{papoutsaki2016webgazer} and mobile tablets and phones~\citep{huang2017tabletgaze,krafka2016eye}.

Gaze data can be collected in parallel when providing other types of feedback, and potentially be combined with previously introduced learning methods. \cite{saran2020efficiently} has shown that incorporating gaze information into imitation from observation (\IfO) and inverse reinforcement learning can lead to a large performance increase in Atari games. Since attention is an intermediate mechanism between perception and action, it becomes very useful when action information is missing in the case of \IfO. In learning evaluative feedback and preference, gaze data might reveal more information to the learning agent to explain why the human gives a particular evaluation. Attention learning is closely related to hierarchical imitation, since gaze is a good indicator of the current high-level behavioral goal which might help an imitator to infer this goal. However, the problem of inferring behavioral goals from human attention needs to be solved first.



\section{Conclusion and Future Directions}
In this survey, we have provided a literature review of progress in leveraging five different types of human guidance (i.e., human inputs that do not involve explicitly defining a reward function or providing an action demonstration) to solve sequential decision-making tasks. In particular, we discussed techniques that have been proposed in the literature that learn from human-provided evaluative feedback, preference, goals, action-free demonstrations, and attention. In each section above, we have discussed future research directions for each approach. Here we briefly discuss several issues and associated potential research questions that are common to all the approaches that leverage human guidance as learning signals. 

\subsection{Shared Datasets and Reproducibility} 
In general, researchers collect their own human guidance data. However, this type of data is often expensive to collect. An effort that could greatly facilitate research in this field is to create publicly available benchmark datasets. Collecting and reusing such datasets may be difficult for some interactive learning methods, in which the guidance (such as evaluative feedback) depends on the changing policy as it is being learned. But, for other approaches, data can be collected in advance and shared. In Table~\ref{tbl:compare} we provide links to existing datasets that are publicly available. Another concern is reproducibility in RL~\citep{henderson2018deep}. When collecting human guidance data, factors such as individual expertise, experimental setup, data collection tools, dataset size, and experimenter bias could introduce large variances in the final performance. Therefore, evaluating algorithms using a standard dataset could save effort and assure a fair comparison between algorithms.

\subsection{Understanding Human Trainers} Leveraging human guidance to train an agent naturally follows a teacher-student paradigm. Much effort has been spent on making the student more intelligent. However, understanding the behavior of human teachers is equally important. \cite{thomaz2008teachable} pioneered the effort in understanding human behavior in teaching learning agents. As RL agents become more powerful and attempt to solve more complex tasks, the human teachers' guiding behaviors could become more complicated and require further study. 

Studying this aspect of human behavior, especially the limitations of human teachers, allows one to design a teaching environment that is more effective and produces more useful guidance data. \cite{amir2016interactive} studied human attention limits while monitoring the learning process of an agent and proposed an algorithm for the human and the agent to jointly identify states where feedback is most needed to reduce human monitoring cost. \cite{ho2016showing} showed the differences in behavior when a human trainer is intentionally teaching (showing) versus merely doing the task. They found that humans modify their policies to reveal the goal to the agent when in the showing mode but not in doing mode. They further showed that imitation learning algorithms can benefit substantially more from the data collected in the showing mode~\citep{ho2016showing}. An important factor to consider is the human trainer's knowledge of the task. \cite{laskey2016robot} have shown that using a hierarchy of human supervisors with different expertise levels can substantially reduce the burden on the experts. 

Understanding the variations in human guidance signals allows algorithms to learn more effectively. We have already seen the debate on how to interpret human evaluative feedback in complex tasks. A helpful way to resolve this debate is to conduct human studies with diverse subject pools to investigate whether real-life human feedback satisfies their algorithmic assumptions and what factors affect the human feedback strategy~\citep{cederborg2015policy,loftin2016learning,macglashan2017interactive}.  

\subsection{An Interactive Paradigm} The best learning results often come from an interactive teaching and learning process in a teacher-student paradigm which involves active
instruction by the human and active learning by the agent. Two factors justify an interactive learning paradigm in our context: \emph{(1)} We only have a partial understanding of human guidance behaviors; and \emph{(2)} the nature of human guidance may vary during training according to the behaviors of the learning agents. As shown by Cooperative IRL~\citep{hadfield2016cooperative}, an iterative and interactive learning process can greatly enhance learning. Therefore the same idea may benefit the process of learning from human guidance as well. 

For evaluative feedback, we have seen a debate on how we interpret human feedback. We have also seen methods that are robust to many types of feedback or that can infer and adapt to different human feedback types~\citep{grizou2014interactive,loftin2016learning,najar2020interactively}. However, perhaps an alternative is to allow the agent to actively query humans for a certain type of feedback that best informs the agent. Learning from human preference is naturally an interactive process when the learning agents actively query for human preferences as we have discussed. Additionally, the aforementioned challenge of selecting optimal trajectory length for query likely requires two-way communication between the human and the agent. In hierarchical imitation, imitation from observation, and attention learning we rarely see examples of interactive learning~\citep{mohseni2015interactive}, since the human data is often collected off-line (see Table~\ref{tbl:compare}) before training as in standard imitation learning. From DAgger~\citep{ross2011reduction} it is clear that interactive training can also benefit imitation learning, however, making the three learning paradigms above interactive remains to be explored.

\subsection{A Unified Learning Framework} The learning frameworks discussed in this paper are often inspired by real-life biological learning scenarios that correspond to different learning stages and strategies in lifelong learning. Imitation and reinforcement learning correspond to learning completely by imitating others and learning completely through self-generated experience, where the former may be used more often in the early stages of learning and the latter could be more useful in the late stages. The other learning strategies discussed are often mixed with these two to allow an agent to utilize signals from all possible sources. For example, it is widely known that children learn largely by imitation and observation~\citep{bandura1961transmission} at their early stage of learning. Then the children gradually learn to develop joint attention with adults through gaze following~\citep{goswami2008cognitive}. Later children begin to adjust their behaviors based on the evaluative feedback and preference received when interacting with other people. Once they developed the ability to reason abstractly about task structure, hierarchical imitation becomes feasible. At the same time, learning through trial and error from reinforcement is always one of the most common types of learning~\citep{skinner1990behavior}. The human's ability to learn from all types of resources continue to develop through a lifetime. We have compared these learning strategies within an imitation and reinforcement learning framework. Under this framework, it is possible to develop a unified learning paradigm that accepts multiple types of human guidance. We start to notice efforts towards this goal~\citep{abel2017agent,waytowich2018cycle,goecks2019,woodward2019learning,najar2020interactively,biyik2020learning}.

In conclusion, the goal of this survey is to serve as a high-level overview of five recent learning frameworks that leverage human guidance to solve sequential decision-making tasks, especially in the context of deep reinforcement learning. We compare and contrast these frameworks by reviewing the motivation, assumption, and implementation of each framework. We hope this will allow researchers in the related areas to see the connections between the works being surveyed, and inspire more research to be done in this field.


\bibliographystyle{spbasic}      

\bibliography{jaamas.bib}

\end{document}